\documentclass[journal, twoside, twocolumn]{IEEEtran}
\ifCLASSOPTIONcompsoc
  \usepackage[nocompress]{cite}

\else
  \usepackage{cite}
\fi
\ifCLASSINFOpdf
\else
\fi
\usepackage{amsmath}
\usepackage{amssymb}
\usepackage{amsmath}
\usepackage{balance}
\usepackage{algorithm}
\usepackage{algorithmic}
\usepackage{multirow}
\usepackage{subfigure}
\usepackage{graphicx}
\usepackage[table,xcdraw]{xcolor}
\usepackage[normalem]{ulem}
\usepackage{subfigure}
\usepackage{hyperref}
\useunder{\uline}{\ul}{}
\usepackage{multirow}

\begin{document}
%
\title{One-step Multi-view Clustering with Diverse Representation}
%
%
%
%

\author{~Xinhang~Wan,~Jiyuan~Liu,~Xinwang~Liu$^{\ast}$,~\IEEEmembership{Senior~Member,~IEEE},~Siwei~Wang,\\~Yi~Wen,~Tianjiao~Wan,~Li~Shen$^{\ast}$,~En~Zhu$^{\ast}$
\IEEEcompsocitemizethanks{
\IEEEcompsocthanksitem X. Wan, J. Liu, X. Liu, S. Wang, Y. Wen, T. Wan, L. Shen, and E. Zhu are with School of Computer, National University of Defense Technology, Changsha, 410073, China. (E-mail: \{wanxinhang,\, liujiyuan13,\, xinwangliu,\, wangsiwei13,\, wenyi,\, wantianjiaocom,\, lishen,\, enzhu\} @nudt.edu.cn).
\IEEEcompsocthanksitem $^{\ast}$: Corresponding author.}
\thanks{Manuscript received Jun. 26, 2023.}
}

\markboth{IEEE Transactions on Neural Networks and Learning Systems}%
{WAN \MakeLowercase{\textit{et al.}}: One-step Multi-view Clustering with Diverse Representation}
\maketitle

\begin{abstract}
Multi-view clustering has attracted broad attention due to its capacity to utilize consistent and complementary information among views. Although tremendous progress has been made recently, most existing methods undergo high complexity, preventing them from being applied to large-scale tasks. Multi-view clustering via matrix factorization is a representative to address this issue. However, most of them map the data matrices into a fixed dimension, limiting the model's expressiveness. Moreover, a range of methods suffers from a two-step process, i.e., multimodal learning and the subsequent $k$-means, inevitably causing a sub-optimal clustering result. In light of this, we propose a one-step multi-view clustering with diverse representation method, which incorporates multi-view learning and $k$-means into a unified framework. Specifically, we first project original data matrices into various latent spaces to attain comprehensive information and auto-weight them in a self-supervised manner. Then we directly use the information matrices under diverse dimensions to obtain consensus discrete clustering labels. The unified work of representation learning and clustering boosts the quality of the final results. Furthermore, we develop an efficient optimization algorithm with proven convergence to solve the resultant problem. Comprehensive experiments on various datasets demonstrate the promising clustering performance of our proposed method.
\end{abstract}

\begin{IEEEkeywords}
Multi-view clustering, Large-scale clustering, Matrix factorization.
\end{IEEEkeywords}


\section{Introduction}\label{sec:introduction}
\begin{figure*}[htbp]
 	\centering
 	\subfigure{
 		\includegraphics[width=0.95\textwidth]{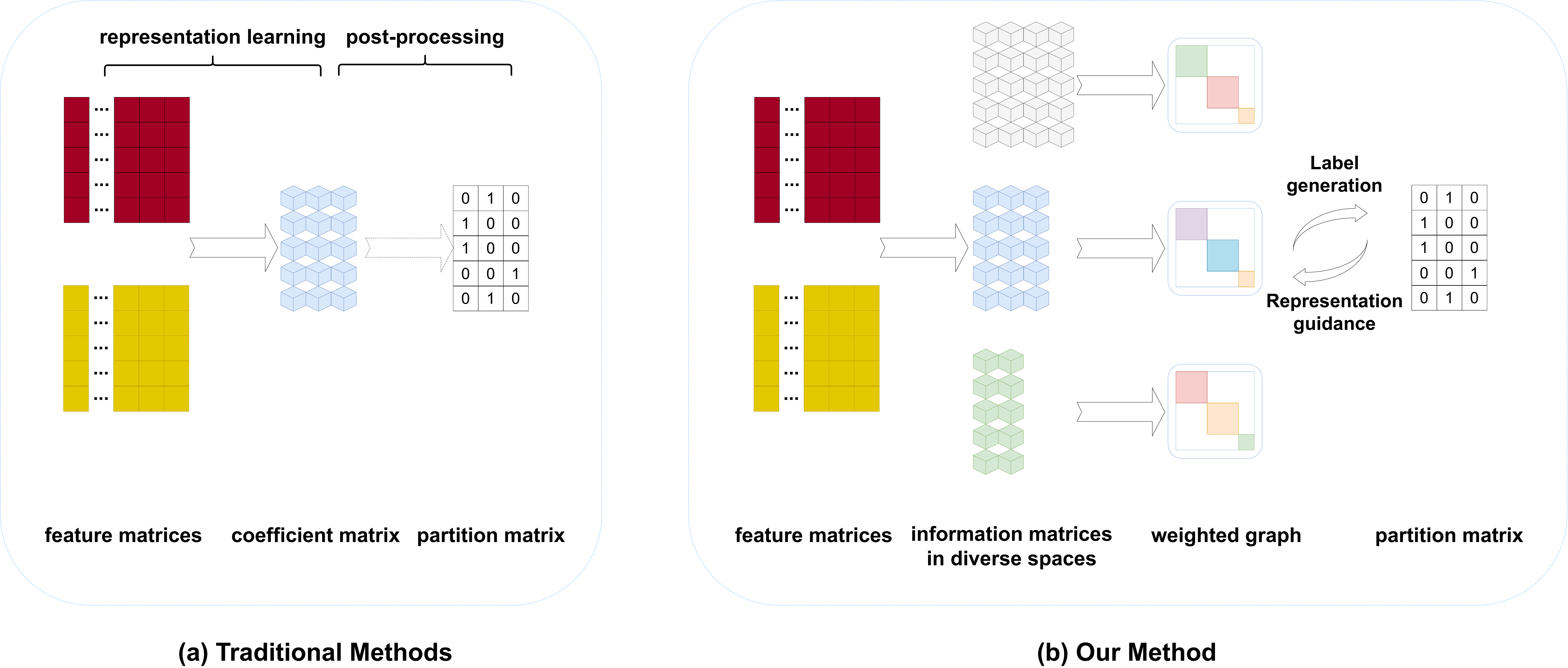}}
 	\caption{The framework comparison between traditional methods (left) and ours (right). Traditional methods conduct representation learning and clustering in two separate steps and fail to select a suitable latent space for each view. On the contrary, our method unifies representation learning and clustering into a unified framework and maps original features into diverse spaces to enhance the expressiveness of the model.
 	}
 	\label{alg_fig}
\end{figure*}

%
%
%
%
\IEEEPARstart{C}{LUSTERING} aims to group samples into several clusters based on the similarities among samples without label information and plays a vital role in machine learning \cite{9063441,9573366,8691702}. However, most clustering algorithms assume that data samples are collected from a single source and neglect multi-view data. In real applications, data frequently comes from multiple sources, and such multi-view data often contains more information \cite{8946880,8440680,9868123}. For example, a user can be described on an online shopping platform by his browser history, shopping cart, consumption records, etc. How to label them is vital for recommending goods personally.

Multi-view clustering, which utilizes consistent and complementary information among views to conduct clustering, has been a popular method to deal with multi-view data \cite{7045574,9860070,5597951,8886709,7902214,7348699}. To the best of our knowledge, existing multi-view clustering algorithms can be roughly divided into two categories, including neural network-based multi-view clustering (NNMVC) and heuristic-based multi-view clustering (HMVC) \cite{10108535}. NNMVC methods attempt to obtain better performance with the feature representation capability of deep models. However, most lack rigorous mathematical formulation and fail to be interpretable \cite{9839616,9258396}. HMVC alleviates it and can be further classified into four types, including multi-view subspace clustering (MVSC) \cite{7410839,9399655,liu2022efficient,8421595,8741173}, multiple kernel clustering (MKC) \cite{9653838, LiangTNNLS, ZJPACMMM, 10.1145/3503161.3547864,9093116}, graph-based methods (MGC) \cite{2020MultiLiang,tang2020,liu-1,liu-3, XGWGCT2022}, and non-negative matrix factorization-based methods (MVCNMF) \cite{8246585,7837980,9170204, Gao2019MultiviewLM}. In specific, MVSC is under a self-expressive framework and assumes that data samples can reconstruct themselves with a reconstruction matrix. For example, \cite{8099944} simultaneously explores the latent representation of data samples and the underlying complementary information to enhance a more comprehensive reconstruction matrix. In MKC \cite{9556554}, most existing methods search a common cluster assignment matrix by jointly maximizing the partition matrix and kernel coefficients. As pointed out by \cite{9146384}, MGC relies on spectral clustering, which performs eigen decomposition upon the Laplacian matrix to partition the data cloud with a linear/nonlinear relationship. Although much attention has been focused on the first three types of algorithms, they often suffer from high time and space complexity, which is unsuitable for large-scale applications. In specific, in MVSC, it takes $\mathcal{O}\left(n^{2}\right)$ to store the subspace structure and $\mathcal{O}\left(n^{3}\right)$ to perform eigen decomposition on it. Similarly, the eigen decomposition process is common in MGC and MKC, preventing them from being applied to large-scale situations.

Owing to the low time and space complexity, multi-view clustering via non-negative matrix factorization method (MVCNMF) is a representative to handle large-scale data and has been surveyed in recent years \cite{10.1016/j.knosys.2019.105185,LIANG2020105582,LUO20181,8387526}. As pointed out by \cite{liu2013multi}, MVCNMF factorizes the original feature matrices into two components, i.e., a consensus coefficient matrix and view-specific base matrices. For instance, \cite{8030316} proposes an algorithm to reduce redundancy by using a diversity term and preserving the local geometry structure of data in each view. Based on the observation that the non-negative constraint guides a less discriminatory embedding, \cite{9710821} removes the non-negative restriction and develops an MF-based algorithm.

However, most existing MVCNMF methods assume that the latent embedding of multiple views is under a fixed dimension. In most cases, the proper latent spaces of different views exist variance, and mapping them into a fixed space is unreasonable. The plain hypothesis harms the expressiveness and the ability to extract complementary information between views. Moreover, most studies fail to directly attain a discrete partition matrix and partition representation learning and clustering into two separate steps. The isolation between them leads to a sub-optimal solution caused by the information loss during the two procedures, which inevitably affects the clustering performance \cite{9769920}. Solving the mentioned drawbacks simultaneously is a tough task, and we summarize them as follows. 1) How to choose a suitable latent space for each view instead of mapping them into a fixed dimension? 2) How to obtain the clustering results directly and combine representation learning and clustering into a unified framework? 3) If we seek a specific latent space for each view, how can we effectively integrate them and avoid the information loss caused by integrating matrices with different dimensions?

To deal with these issues, we propose a one-step multi-view clustering with diverse representation (OMVCDR) method. The framework of our proposed method is shown in Fig. \ref{alg_fig}. To enhance the data embedding and gain complementary information among views, we obtain various information matrices learned by base matrices under diverse dimensions and auto-weight them according to their contributions to the final result. Although comprehensive knowledge benefits the final result, it brings a new task for exploiting it. Most conventional methods address it by imposing a consensus matrix to match them and conduct $k$-means on it. However, the information loss happens when we align matrices of different dimensions and isolate representation learning with $k$-means, leading to poor clustering performance. In light of this, we directly utilize the information matrices to get final clustering assignments by calculating the distances among samples and performing $k$-means on them. In this way, the various information matrices can induce a partition matrix with high quality. As feedback, the consensus one guides the generation of information matrices likewise. We introduce a hyper-parameter to balance the weights between the representation learning and the clustering process. Given that the distance matrices might increase our model's time and space complexity, we design an efficient algorithm to calculate the distances of samples in the same cluster. All in all, the primary contributions of OMVCDR are summarized as follows:

 \begin{enumerate}
\item By directly calculating the distance among samples with diverse representation, we incorporate MVCMF and $k$-means into a unified framework with linear complexity. They negotiate with each other and boost the clustering performance. 
\item The information loss is lessened in three aspects. First, we get information matrices in diverse latent spaces to enhance adequate knowledge. Second, the matrices are directly used to attain final results, and we avoid the alignment of matrices under different dimensions. Third, the separation between representation learning and clustering is overcome.
\item We develop a four-step alternative optimization algorithm with proven convergence. To validate the effectiveness of OMVCDR, we conduct extensive experiments on various datasets. The results demonstrate the efficiency and excellent performance of the proposed method.
\end{enumerate}

This work is a substantially extended version of our original conference paper \cite{wanAAAI2023}. Its significant improvement over the previous one can be summarized as follows: 1) We design a new method, termed OMVCDR, by incorporating diverse representation learning and clustering into a unified framework, and develop an alternating optimization algorithm to solve the resultant problem. Furthermore, the clustering performance comparison demonstrates that OMVCDR is significantly superior to AWMVC proposed in the previous paper. 2) Compared with AWMVC, the information loss is lessened in two areas, i.e., the match of matrices of different dimensions is avoided, and the separation between representation learning and clustering is overcome. 3) More theoretical analysis is included. For instance, we build the bridge between our work and anchor learning. 4) Different from traditional one-pass algorithms with quadratic or cubic time complexity \cite{9769920,9298842}, we design an efficient distance calculation strategy with linear complexity.
5) Besides detailed extension and clarification, we perform comprehensive experiments to verify the effectiveness of our proposed method. Although our proposed method seems simple, the attempt to unify diverse representation learning and clustering into a process with linear complexity might inspire other research, such as anchor learning.

The rest of the paper is organized as follows. Section 2 introduces the related work. In Section 3, we give the motivation and formulation of OMVCDR, then we design an alternating algorithm to solve the resultant problem. Section 4 provides theoretical analysis, including the convergence proof, time and space complexity, and the connection between OMVCDR and anchor learning. We conduct extensive experiments to verify the excellent performance of our proposed method in Section 5. Section 6 shows the conclusions.

\section{related work}

In this section, we briefly overview the related work concerning $k$-means, single-view clustering via matrix factorization (SVCMF), and multi-view clustering via matrix factorization (MVCMF).
\subsection{\texorpdfstring{$K$}--means}

As a traditional machine learning algorithm, $k$-means partitions data samples into $k$ groups by maximizing the similarity among samples in the same cluster, where $k$ is a pre-defined value. Specifically, \cite{2007The} proposes the loss function of typical $k$-means as follows,
\begin{equation}\label{kmeans}
\begin{aligned}
\sum_{j=1}^k \sum_{i \in C_j}\left\|\mathbf{x}_i-\mathbf{\mu}_j\right\|_2^2,
\end{aligned}
\end{equation}
where $\mathbf{x}_i \in \mathbb{R}^{1 \times d}$ and $\mathbf{\mu}_j\in \mathbb{R}^{1 \times d}$ denote the feature vector corresponding to $i$-th sample and the centroid in $j$-th cluster, respectively.

Under the hypothesis that each cluster shares a balanced number of samples, Pei \textit{et al.} \cite{PMID:35157578} develop a novel $k$-means, shown as,
\begin{equation}\label{ksumx}
\begin{aligned}
\min _{\mathbf{Y} \in \Phi^{n \times k}} \operatorname{Tr}\left(\mathbf{Y}^T \mathbf{D Y}\right),
\end{aligned}
\end{equation}
where $\mathbf{Y} \in \Phi^{n \times k}$ is a cluster partition matrix and $\mathbf{D}\in \mathbb{R}^{n \times n}$ is the distance matrix where $d_{ij}=\left\|\mathbf{x}_i-\mathbf{x}_j\right\|_2^2$. Furthermore, they propose an accelerated algorithm to solve the problem. However, this method fails to handle multi-view data since it assumes the data is obtained from a single source.
\subsection{SVCMF}
Given a single-view dataset $\mathbf{X}=\left[\mathbf{x}_{1}, \ldots, \mathbf{x}_{n}\right] \in \mathrm{R}^{d \times n}$, where $d$ and $n$ represent feature dimension and sample number separately, SVCNMF factorizes feature matrix $\mathbf{X}$ into two non-negative components, i.e., the base matrix $\mathbf{H}\in \mathrm{R}^{d \times k}$ and coefficient matrix $\mathbf{Z}\in \mathrm{R}^{k \times n}$, where $k$ is the cluster number. SVCNMF aims to approximate the product of $\mathbf{H}$ and $\mathbf{Z}$ to $\mathbf{X}$ as follows,

 \begin{equation}\label{ORIGINAL_NMFSVC}
\begin{aligned}
\min _{\mathbf{H} \geq \mathbf{0}, \mathbf{Z} \geq \mathbf{0}} f(\mathbf{X}, \mathbf{H Z}),
\end{aligned}
\end{equation}
where $f(\cdot)$ is a loss function. Most existing methods \cite{liu2013multi,9163943} adopt Frobenius norm to measure the loss of representation learning. Given that the orthogonality constraint on NMF guides a rigorous clustering interpretation \cite{2006Orthogonal}, a series of methods have been proposed under the framework of orthogonality constraint. By imposing it on base matrices, Eq. \eqref{ORIGINAL_NMFSVC} can be redefined as,
\begin{equation}\label{ORTHO_NMFSVC}
\begin{aligned}
\min _{\mathbf{H} \geq \mathbf{0}, \mathbf{Z} \geq \mathbf{0}}\|\mathbf{X}-\mathbf{H Z}\|_{F}^{2} \quad \text { s.t. } \mathbf{H}^{\top} \mathbf{H}=\mathbf{I}_{k}.
\end{aligned}
\end{equation}

Based on the observation that the non-negative constraint leads to less discriminatory information, \cite{9710821} removes the restriction and develops an SVCMF method, shown as,
\begin{equation}\label{ORTHO_MFSVC}
\begin{aligned}
\min _{\mathbf{H}, \mathbf{Z}}\|\mathbf{X}-\mathbf{H Z}\|_{F}^{2} \quad \text { s.t. } \mathbf{H}^{\top} \mathbf{H}=\mathbf{I}_{k}.
\end{aligned}
\end{equation}

After obtaining the coefficient matrix $\mathbf{Z}$, a subsequent clustering method will be performed on it to attain the final cluster assignments.
\subsection{MVCMF}

Given a multi-view data $\left\{\mathbf{X}^{(v)}\right\}_{v=1}^{V}$, where $\mathbf{X}^{(v)} \in \mathbf{R}^{d_{v} \times n}$ and $d_{v}$ represents the feature dimension of $v$-th view, MVCMF seeks to attain a consensus matrix $\mathbf{Z}$ containing abundant information. The typical formulation is as follows \cite{8030316}, 
\begin{equation}\label{ORI_MFMVC}
\begin{aligned}
&\min _{\left\{\mathbf{H}^{(v)}\right\}, \left\{\mathbf{Z}^{(v)}\right\}}\sum_{v=1}^{V}\frac{1}{2}\left\|\mathbf{X}^{(v)}-\mathbf{H}^{(v)} \mathbf{Z}^{(v)}\right\|_{F}^{2}\\
&+\lambda \sum_{v \neq w} \operatorname{DIVE}\left(\mathbf{Z}^{(v)}, \mathbf{Z}^{(w)}\right) \\
&\text { s.t. }  1 \leq v, w \leq V, \mathbf{H}^{(v)}\geq \mathbf{0}, \mathbf{Z}^{(v)}\geq \mathbf{0}, \mathbf{Z}^{(w)} \geq \mathbf{0},
\end{aligned}
\end{equation}
where $\operatorname{DIVE}\left(\cdot,\cdot\right)$ is a penalty term to reduce the redundancy among representation matrices and $\lambda$ is the balanced hyper-parameter. The consensus matrix $\mathbf{Z}$ is gained via the average value of $\mathbf{Z}^{(v)}$. On the contrary, some algorithms directly get a common underlying matrix across views, shown as follows,

\begin{equation}\label{ORTHO_MFMVC}
\begin{aligned}
&\min _{\left\{\mathbf{H}^{(v)}\right\}, \mathbf{Z}} \sum_{v=1}^{V}\frac{1}{2}\left\|\mathbf{X}^{(v)}-\mathbf{H}^{(v)} \mathbf{Z}\right\|_{F}^{2}+\lambda\Phi(\mathbf{H}^{(v)}, \mathbf{Z}) \\
&\text { s.t. } {\mathbf{H}^{(v)}}^{\top} \mathbf{H}^{(v)}=\mathbf{I}_{k},
\end{aligned}
\end{equation}
where $\Phi(\cdot)$ are some regularization terms on $\mathbf{H}$ and $\mathbf{Z}$. 

Although these methods have fulfilled MVCMF from different aspects, they still encounter the following drawbacks: 1) The representation ability is limited since existing methods map data samples of different views into a specific latent space. However, in real applications, the suitable spaces are frequently under different dimensions, and projecting them into a fixed dimension is pernicious. 2) The representation learning and clustering are isolated. The separation between the two processes leads to a sub-optimal solution since the information loss generates during the two steps. To handle these shortcomings, we propose a novel MVCMF algorithm.

\section{One-step Multi-view Clustering with Diverse Representation}

In this section, we first describe the motivation and formulation of our proposed method. Then we provide an alternating optimization algorithm to solve the resultant problem.
\subsection{Motivation and Proposed Formula}
As analyzed above, existing MVCMF generally drops in two deficiencies, i.e., limited representation ability and the isolation between representation learning and clustering. However, solving them simultaneously is an arduous task. The main challenges are as follows: 1) How to enhance the representation ability to ensure the quality of representation learning? 2) How to utilize the information matrices effectively and avoid the information loss between representation learning and clustering?

In our previous work \cite{wanAAAI2023}, we solved the first challenge by learning coefficient matrices in diverse latent spaces and auto-weighting them according to their contributions to the final result. Despite strengthening the expressiveness, the model fails to connect representation learning with clustering. Additionally, the alignment of the matrices under different dimensions also results in the loss of knowledge.

To address these concerns, we propose one-step multi-view clustering with diverse representation (OMVCDR). Different from AWMVC, we directly conduct $k$-means with the information matrices conveying comprehensive knowledge. Specifically, we construct weighted graphs using the outcome of representation learning and seek a consensus hard partition matrix among them. The information loss is reduced since we straightforwardly conduct clustering on the information matrices rather than matching them with a consensus matrix and performing $k$-means subsequently. In this way, comprehensive knowledge can induce a high-quality consensus partition matrix. As feedback, the latter guides the generation of information matrices likewise. A hyper-parameter is introduced to balance the weight between representation learning and clustering. The optimization goal is as follows,
\begin{equation}\label{OMVCDR}
\begin{aligned}
&\min _{\substack{\left\{\mathbf{Z}_p\right\}, \left\{\mathbf{H}_p^{(v)}\right\}, \\ \mathbf{Y} \in \Phi^{n \times k}, \boldsymbol{\alpha}}} \sum_{p=1}^m \sum_{v=1}^V \left\|\mathbf{X}^{(v)}-\mathbf{H}_p^{(v)} \mathbf{Z}_p\right\|_F^2\\
&+\frac{\lambda}{2}  \operatorname{Tr}\left(\mathbf{Y}^{\top}\left(\sum\nolimits_{p=1}^m \alpha_p^2 \mathbf{D}_p\right) \mathbf{Y}\right)\\
&\text { s.t. } {{\mathbf{H}_p^{(v)}}^{\top} \mathbf{H}_p^{(v)}=\mathbf{I}_{d_p}, \boldsymbol{\alpha}^{\top} \mathbf{1}=1, \boldsymbol{\alpha}\geq\mathbf{0},}
\end{aligned}
\end{equation}
where $\mathbf{H}_p^{(v)}\in\mathbb{R}^{d_v \times d_p}$ is the basic matrix of $v$-th view under $p$-th dimension, $\mathbf{Z}_p$ denotes the consensus information matrix of $p$-th latent space, $\mathbf{Y}$ represents the final clustering result, and $\mathbf{D}_p\in\mathbb{R}^{n \times n}$ is computed by $d_p^{ij}=\left\|\mathbf{z}_p^{i}-\mathbf{z}_p^{j}\right\|_2^2$.

From the formula, it is obtained that the information loss is reduced in three areas: 1) The diverse latent spaces compensate for inadequate feature learning. 2) The information matrices are directly utilized instead of matching them with a low-dimensional consensus matrix. 3) The isolation between representation learning and clustering is avoided. Meanwhile, the information matrices and partition matrix negotiate with each other to boost the clustering performance. Apparently, the generation and storage of the distance matrices is quadratic .
complexity respecting the sample number. In the optimization process, we provide an efficient strategy to handle it with linear complexity.
\subsection{Optimization}   
There are four variables in Eq. (\ref{OMVCDR}) and optimizing them simultaneously is challenging. Therefore, we provide an alternating algorithm to update one variable while the others are fixed.

\subsubsection{\texorpdfstring{$\mathbf{Z}_p$} -Subproblem}
Let $\mathbf{Z}_p=[\mathbf{z}_p^{1}, \mathbf{z}_p^{2}, ..., \mathbf{z}_p^{n}]$ and fix other variables, Eq. (\ref{OMVCDR}) can be rewritten as,
\begin{equation}\label{opt_Z}
\begin{aligned}
&\min _{\mathbf{Z}_p}  \sum_{v=1}^V \operatorname{Tr}\left(\mathbf{Z}_p^{\top} \mathbf{Z}_p-2{\mathbf{X}^{(v)}}^{\top} \mathbf{H}_{p}^{(v)}\mathbf{Z}_p\right)\\
&+\lambda\alpha_p^2 \sum_{c=1}^k \sum_{i, j \in \mathcal{A}_c}\left\|\mathbf{z}_p^{i}-\mathbf{z}_p^{j}\right\|_2^2,
\end{aligned}
\end{equation}
where $i, j \in \mathcal{A}_c$ represents that sample $i$ and $j$ are in the $c$-th cluster.

Considering that each column of $\mathbf{Z}_p$ is coupled with each other, optimizing $\mathbf{Z}_p$ is tough, and we optimize each column of it separately as,
\begin{equation}
\begin{aligned}
&\min _{\mathbf{z}_p^i}  \sum_{v=1}^V \left({\mathbf{z}_p^i}^{\top} \mathbf{z}_p^i-2{(\mathbf{X}^{(v)}}^{\top} \mathbf{H}_{p}^{(v)})(i,:)\mathbf{z}_p^i\right)\\
&+\lambda\alpha_p^2 \sum_{j\in\mathcal{A}_{c_i},j\neq i }\left\|\mathbf{z}_p^{i}-\mathbf{z}_p^{j}\right\|_2^2.
\end{aligned}
\end{equation}

Setting the derivative to zero, it is obtained that $\mathbf{z}_p^i$ can be updated via
\begin{equation}\label{get_zpi}
\begin{aligned}
\mathbf{z}_{p}^i=(\mathbf{a}_{p}^i+\lambda\alpha_p^2\sum_{j\in\mathcal{A}_{c_i},j\neq i }\mathbf{z}_p^{j}) /(V+\lambda\alpha_p^2(|\mathcal{A}_{c_i}|-1)),
\end{aligned}
\end{equation}
where $\mathbf{A}_p=\sum_{v=1}^V{\mathbf{H}_{p}^{(v)}}^{\top}\mathbf{X}^{(v)}$ and $\mathcal{A}_{c_i}$ is the set of samples in the same cluster as the $i$-th sample.

\subsubsection{\texorpdfstring{$\mathbf{H}_{p}^{(v)}$}- Subproblem}
Given that the base matrices are independent of each other, we provide the optimization process of the base matrix $\mathbf{H}_{p}^{(v)}$ of $v$-th view under $p$-th dimension as an example. When other variables maintain fixed, the optimization problem in Eq. \eqref{OMVCDR} is equivalent to,
\begin{equation}\label{opt_H}
\begin{aligned}
\max_{\mathbf{H}_p^{(v)}} \operatorname{Tr}\left({\mathbf{H}_p^{(v)}}^{\top} \mathbf{B}_p^{(v)}\right) \text { s.t. }
{\mathbf{H}_p^{(v)}}^{\top} \mathbf{H}_p^{(v)}=\mathbf{I}_{d_p},
\end{aligned}
\end{equation}
where $\mathbf{B}_p^{(v)}=\mathbf{X}^{(v)} \mathbf{Z}_p^{\top}$.

Suppose that the matrix $\mathbf{B}_p^{(v)}$ has the singular value decomposition (SVD) form as $\mathbf{B}_p^{(v)}=\mathbf{S}_p^{(v)} \boldsymbol{\Sigma}_p^{(v)} {\mathbf{V}_p^{(v)}}^{\mathrm{T}}$, the optimization problem in Eq. (\ref{opt_H}) can be solved by a closed-form solution in \cite{ijcai2019-524} as follows,
\begin{equation}\label{get_hpv}
\begin{aligned}
\mathbf{H}_{p}^{(v)}=\mathbf{S}_p^{(v)}  {\mathbf{V}_p^{(v)}}^{\top}.
\end{aligned}
\end{equation}

\subsubsection{\texorpdfstring{$\mathbf{Y}$} -Subproblem}
Fixing other variables, the objective formula of Eq. \eqref{OMVCDR} is reduced to,
\begin{equation}\label{opt_Y}
\begin{aligned}
\min _{ \mathbf{Y} \in \Phi^{n \times k}} \operatorname{Tr}\left(\mathbf{Y}^{\top}\left(\sum\nolimits_{p=1}^m \alpha_p^2 \mathbf{D}_p\right) \mathbf{Y}\right).
\end{aligned}
\end{equation}

The optimal solution of $\mathbf{y}_i$ can be attained via
\begin{equation}\label{yil}
\begin{aligned}
y_{i l}=\left\{\begin{array}{cc}
1 & l=\arg \min _c\left(\mathbf{m}_i^{\top} \mathbf{Y}\right)_c \\
0 & \text { otherwise, }
\end{array}\right.
\end{aligned}
\end{equation}
where $\mathbf{M}=\sum_{p=1}^m \alpha_p^2 \mathbf{D}_p$.

However, the storage and processing of $\mathbf{M}$ take $\mathcal{O}\left(n^{2}\right)$ complexity at least, which is impractical for large-scale data. Motivated by \cite{PMID:35157578}, we provide an efficient updating algorithm to solve it. 

Indeed, for each $\left(\mathbf{m}_i^{\top} \mathbf{Y}\right)_c$, we have
\begin{equation}\label{mY}
\begin{aligned}
\left(\mathbf{m}_i^{\top} \mathbf{Y}\right)_c=\sum_{p=1}^m \alpha_p^2\sum_{j\in\mathcal{A}_{c}}\left\|\mathbf{z}_p^{i}-\mathbf{z}_p^{j}\right\|_2^2.
\end{aligned}
\end{equation}

The more straightforward formula of the sum of these distances can be written as,
\begin{equation}\label{sum}
\begin{aligned}
\sum_{j\in\mathcal{A}_{c}}\left\|\mathbf{z}_p^{i}-\mathbf{z}_p^{j}\right\|_2^2=|\mathcal{A}_{c}|\left\|\mathbf{z}_p^{i}\right\|_2^2-2{\mathbf{z}_p^{i}}^{\top}\sum_{j\in\mathcal{A}_{c}}\mathbf{z}_p^{j}+\sum_{j\in\mathcal{A}_{c}}\left\|\mathbf{z}_p^{j}\right\|_2^2.
\end{aligned}
\end{equation}

The sum of $\sum_{j\in\mathcal{A}_{c}}\mathbf{z}_p^{j}$ and $\sum_{j\in\mathcal{A}_{c}}\left\|\mathbf{z}_p^{j}\right\|_2^2$ can be pre-calculated when updating $\mathbf{Z}_p$. Therefore, the time complexity of optimizing $\mathbf{Y}$ is linear to the sample number and storing $\mathbf{D}_p$ is unnecessary.

\subsubsection{\texorpdfstring{$\boldsymbol{\alpha}$}- Subproblem}
With other variables fixed, the formulation respecting $\boldsymbol{\alpha}$ can be solved via optimizing the following formula,

\begin{equation}\label{opt_alpha}
\begin{aligned}
\min \sum_{p=1}^{m} \alpha_p^2 r_p^2 \text { s.t. } \boldsymbol{\alpha}^{\top} \mathbf{1}=1, \boldsymbol{\alpha}\geq\mathbf{0},
\end{aligned}
\end{equation}
where
\begin{equation}
\begin{aligned}
r_p^2=\operatorname{Tr}\left(\mathbf{Y}^T \mathbf{D}_p \mathbf{Y}\right).
\end{aligned}
\end{equation}

Based on Cauchy-Schwarz inequality, we have
\begin{equation}
\begin{aligned}
\left(\sum_{p=1}^{m} \alpha_p^2 r_p^2\right) \left(\sum_{p=1}^{m}\frac{1}{r_{p}^2}\right)\geq \sum_{p=1}^{m}\alpha_p=1
\end{aligned}
\end{equation}
in which the equality holds when
\begin{equation}\label{get_alpha}
\begin{aligned}
\alpha_{p}=\frac{\frac{1}{r_{p}^2}}{\sum_{p=1}^{m} \frac{1}{r_{p}^2}}.
\end{aligned}
\end{equation}

For convenience, let $\mathbf{t}_{p}^c=\sum_{j\in\mathcal{A}_{c}}\mathbf{z}_p^{j}$ and $v_{p}^c=\sum_{j\in\mathcal{A}_{c}}\left\|\mathbf{z}_p^{j}\right\|_2^2$, the overview of the alternate optimization strategy is outlined in Algorithm 1.

\begin{algorithm}[h]
\renewcommand{\algorithmicrequire}{\textbf{Input:}}
	\renewcommand{\algorithmicensure}{\textbf{Output:}}
		\caption{One-step Multi-view Clustering with Diverse Representation}
		\label{algo}
		\begin{algorithmic}[1]
			\REQUIRE Dataset $\left\{\mathbf{X}^{(v)}\right\}_{v=1}^{V}$, cluster number $k$ and hyper-parameter $\lambda$.
			\ENSURE Cluster partition matrix $\mathbf{Y}$.
			\STATE Initialize  $\left\{\mathbf{H}_p^{(v)}\right\}$, $\boldsymbol{\alpha}$, $\mathbf{Y} \in \Phi^{n \times k}$, $\left\{\mathbf{T}_p\right\}$ and $\left\{\mathbf{v}_p\right\}$.
   \WHILE{not converged}
   \FOR{$p$ = 1 : $m$}
   \FOR{$i$ = 1 : $n$}
			\STATE Compute $\mathbf{z}_{p}^i$ by  Eq. \eqref{get_zpi}, update $\mathbf{t}_{p}^{c_i}$ and $v_{p}^{c_i}$.
		   \ENDFOR
     \ENDFOR
     \STATE Calculate $\left\{\mathbf{H}_p^{(v)}\right\}$ by Eq. \eqref{get_hpv}.
     \FOR{$i$ = 1 : $n$}
     \STATE Remove $i$-th sample from its cluster.
   \FOR{$p$ = 1 : $m$}
			\STATE Update $\mathbf{t}_{p}^{c_i}$ and $v_{p}^{c_i}$.
		   \ENDFOR
     \STATE Calculate $\mathbf{m}_i^{\top} \mathbf{Y}$ via Eq. \eqref{mY} and \eqref{sum}, then update $\mathbf{y}_i$ by Eq. \eqref{yil}.
     \FOR{$p$ = 1 : $m$}
			\STATE Update $\mathbf{t}_{p}^{c_i}$ and $v_{p}^{c_i}$.
		   \ENDFOR
     \ENDFOR
     \STATE Calculate $\boldsymbol{\alpha}$ by Eq. \eqref{get_alpha}.
     \ENDWHILE
     
		\end{algorithmic}
	\end{algorithm}

\section{ Theoretical Analysis}
\subsection{Convergence}\label{convergence_section}
Most MVCMF methods fail to prove the convergence theoretically, such as \cite{ZONG201774,10.1007/s00530-022-00905-x}, while we verify the convergence of OMVCDR in this section. For ease of expression, we simplify the objective formula in Eq. \eqref{OMVCDR} as,
\begin{equation}
\begin{aligned}
\min_{\left\{\mathbf{Z}_p\right\},\left\{\mathbf{H}_p^{(v)}\right\}, \mathbf{Y} , \boldsymbol{\alpha}} \mathcal{J}\left(\left\{\mathbf{Z}_p\right\},\left\{\mathbf{H}_p^{(v)}\right\}, \mathbf{Y} , \boldsymbol{\alpha}\right).
\end{aligned}
\end{equation}

As shown in Algorithm 1, the optimization process consists of four alternative parts, i.e., $\left\{\mathbf{Z}_p\right\}$, $\left\{\mathbf{H}_p^{(v)}\right\}$, $\mathbf{Y}$, and $\boldsymbol{\alpha}$ subproblems. Assume that the superscript $t$ denotes the optimization at round $t$, the convergence of each subproblem is given as follows:

\subsubsection{\texorpdfstring{$\left\{\mathbf{Z}_p\right\}$}- Subproblem} Given $\left\{\mathbf{H}_p^{(v)}\right\}^{(t)}$, $\mathbf{Y}^{(t)}$ and $\boldsymbol{\alpha}^{(t)}$, $\left\{\mathbf{Z}_p\right\}^{(t+1)}$ can be optained via Eq. \eqref{get_zpi}, resulting in
\begin{equation}\label{leq1}
\begin{aligned}
\mathcal{J}\left(\left\{\mathbf{Z}_p\right\}^{(t+1)},\left\{\mathbf{H}_p^{(v)}\right\}^{(t)}, \mathbf{Y}^{(t)} , \boldsymbol{\alpha}^{(t)}\right)\leq\\
\mathcal{J}\left(\left\{\mathbf{Z}_p\right\}^{(t)},\left\{\mathbf{H}_p^{(v)}\right\}^{(t)}, \mathbf{Y}^{(t)} , \boldsymbol{\alpha}^{(t)}\right),
\end{aligned}
\end{equation}
which indicates the objective value decreases along with  the optimization of $\left\{\mathbf{Z}_p\right\}$.

\subsubsection{\texorpdfstring{$\left\{\mathbf{H}_p^{(v)}\right\}$}-Subproblem}

Given $\left\{\mathbf{Z}_p\right\}^{(t+1)}$, $\mathbf{Y}^{(t)}$ and $\boldsymbol{\alpha}^{(t)}$, $\left\{\mathbf{H}_p^{(v)}\right\}^{(t+1)}$ can be optained via Eq. \eqref{get_hpv}, resulting in
\begin{equation}\label{leq2}
\begin{aligned}
\mathcal{J}\left(\left\{\mathbf{Z}_p\right\}^{(t+1)},\left\{\mathbf{H}_p^{(v)}\right\}^{(t+1)}, \mathbf{Y}^{(t)} , \boldsymbol{\alpha}^{(t)}\right)\leq\\
\mathcal{J}\left(\left\{\mathbf{Z}_p\right\}^{(t+1)},\left\{\mathbf{H}_p^{(v)}\right\}^{(t)}, \mathbf{Y}^{(t)} , \boldsymbol{\alpha}^{(t)}\right),
\end{aligned}
\end{equation}
which indicates the objective value decreases along with  the optimization of $\left\{\mathbf{H}_p^{(v)}\right\}$.

\subsubsection{\texorpdfstring{$\mathbf{Y}$}- Subproblem}

Given $\left\{\mathbf{Z}_p\right\}^{(t+1)}$, $\left\{\mathbf{H}_p^{(v)}\right\}^{(t+1)}$ and $\boldsymbol{\alpha}^{(t)}$, $\mathbf{Y}^{(t+1)}$ can be optained via Eq. \eqref{yil}, resulting in
\begin{equation}\label{leq3}
\begin{aligned}
\mathcal{J}\left(\left\{\mathbf{Z}_p\right\}^{(t+1)},\left\{\mathbf{H}_p^{(v)}\right\}^{(t+1)}, \mathbf{Y}^{(t+1)} , \boldsymbol{\alpha}^{(t)}\right)\leq\\
\mathcal{J}\left(\left\{\mathbf{Z}_p\right\}^{(t+1)},\left\{\mathbf{H}_p^{(v)}\right\}^{(t+1)}, \mathbf{Y}^{(t)} , \boldsymbol{\alpha}^{(t)}\right),
\end{aligned}
\end{equation}
which indicates the objective value decreases along with  the optimization of $\mathbf{Y}$.

\subsubsection{\texorpdfstring{$\boldsymbol{\alpha}$}-Subproblem}

Given $\left\{\mathbf{Z}_p\right\}^{(t+1)}$, $\left\{\mathbf{H}_p^{(v)}\right\}^{(t+1)}$ and $\mathbf{Y}^{(t+1)}$,  $\boldsymbol{\alpha}^{(t+1)}$can be optained via Eq. \eqref{get_alpha}, resulting in
\begin{equation}\label{leq4}
\begin{aligned}
\mathcal{J}\left(\left\{\mathbf{Z}_p\right\}^{(t+1)},\left\{\mathbf{H}_p^{(v)}\right\}^{(t+1)}, \mathbf{Y}^{(t+1)} , \boldsymbol{\alpha}^{(t+1)}\right)\leq\\
\mathcal{J}\left(\left\{\mathbf{Z}_p\right\}^{(t+1)},\left\{\mathbf{H}_p^{(v)}\right\}^{(t+1)}, \mathbf{Y}^{(t+1)} , \boldsymbol{\alpha}^{(t)}\right).
\end{aligned}
\end{equation}
which indicates the objective value decreases along with  the optimization of $\boldsymbol{\alpha}$.

To sum up Eq. \eqref{leq1} to \eqref{leq4}, it is obtained that:
\begin{equation}\label{leq5}
\begin{aligned}
\mathcal{J}\left(\left\{\mathbf{Z}_p\right\}^{(t+1)},\left\{\mathbf{H}_p^{(v)}\right\}^{(t+1)}, \mathbf{Y}^{(t+1)} , \boldsymbol{\alpha}^{(t+1)}\right)\leq\\
\mathcal{J}\left(\left\{\mathbf{Z}_p\right\}^{(t)},\left\{\mathbf{H}_p^{(v)}\right\}^{(t)}, \mathbf{Y}^{(t)} , \boldsymbol{\alpha}^{(t)}\right),
\end{aligned}
\end{equation}
which indicates the objective value monotonically decreases along with iterations.

Given that $ \mathbf{Y} \in \Phi^{n \times k}$, $\alpha_p\geq0$ and $d_p^{ij}\geq0$, it is easy to conclude that:
\begin{equation}
\begin{aligned}
\operatorname{Tr}\left(\mathbf{Y}^{\top}\left(\sum_{p=1}^m \alpha_p^2 \mathbf{D}_p\right) \mathbf{Y}\right)\geq0.
\end{aligned}
\end{equation}

Therefore, the following inequality holds:
\begin{equation}
\begin{aligned}
\mathcal{J}\left(\left\{\mathbf{Z}_p\right\},\left\{\mathbf{H}_p^{(v)}\right\}, \mathbf{Y} , \boldsymbol{\alpha}\right)\geq0.
\end{aligned}
\end{equation}

Thus, the objective formula exists a lower bound and the proposed method is convergent theoretically.

\subsection{Time Complexity}
The optimization process comprises four subproblems, and we analyze the time complexity of each subproblem separately per iteration. In $\left\{\mathbf{Z}_p\right\}$ subproblem, it costs $\mathcal{O}\left(dd_pn\right)$ to get $\mathbf{A}_p$, where $d=\sum_{v=1}^V d_v$. Then the time complexity of computing $\mathbf{Z}_p$ is $\mathcal{O}\left(d_pn\right)$. Thus, it costs $\mathcal{O}\left(d\sum_{p=1}^md_pn\right)$ to update $\left\{\mathbf{Z}_p\right\}$. In $\left\{\mathbf{H}_p^{(v)}\right\}$ subproblem, it takes $\mathcal{O}\left(d_vd_pn\right)$ to calculate $\mathbf{B}_p^{(v)}$ and $\mathcal{O}\left(d_vd_p^2\right)$ to perform SVD on it. Thus, it costs $\mathcal{O}\left(d\sum_{p=1}^md_pn+\sum_{p=1}^mdd_p^2\right)$ to update $\left\{\mathbf{H}_p^{(v)}\right\}$. In $\mathbf{Y}$ subproblem, since $\mathbf{t}_{p}^{c}$ and $v_{p}^{c}$ are pre-calculated, it requires $\mathcal{O}\left(mkn\right)$ to update $\mathbf{Y}$. In $\boldsymbol{\alpha}$ subproblem, it takes $\mathcal{O}\left(1\right)$ to update it. Assume that the algorithm converges in $T$ iterations, the total complexity is $\mathcal{O}\left(Td\sum_{p=1}^md_pn+T\sum_{p=1}^mdd_p^2\right)$. Consequently, the proposed optimization algorithm is linear complexity respecting the sample number.

\subsection{Space Complexity}
The primary memory costs of our proposed method are the storage of matrices $\mathbf{H}_p^{(v)}\in\mathbb{R}^{d_v \times d_p}$, $\mathbf{Z}_p\in\mathbb{R}^{d_p \times n}$ and $\mathbf{Y}\in\mathbb{R}^{n \times k}$. Therefore, the total space complexity of our proposed method is $\mathcal{O}\left((d+n)\sum_{p=1}^md_p\right)$, which is linear complexity respecting the sample number as well.

%
\subsection{Connection with Anchor Learning}
Multi-view clustering based on anchor learning seeks to establish an anchor graph to avoid the high complexity of graph learning on the original graph \cite{9180083}. It is easy to show that it is a special case of our work when $\mathbf{H}_p^{(v)}$ and $\mathbf{Z}_p$ are considered as the anchor representation and anchor graph, respectively. In anchor learning, the involved anchor number selection is hard to choose since it lacks label information in a clustering task \cite{10.1145/3474085.3475516}. Therefore, our research might inspire corresponding work and provide a manner to automatically choose a suitable anchor number for each view and directly utilize the anchor graphs with less information loss.

\section{Experiment}
In this section, we evaluate the efficiency and effectiveness of OMVCDR on seven widely used datasets in terms of clustering performance and running time. After that, we conduct an ablation study to verify the validity of the critical components of our proposed method, then analyze the convergence and parameter sensitivity of it.
\subsection{Experimental Settings}
\subsubsection{Dataset}
Seven widely used datasets are employed to verify the superiority of our proposed method, including Flower17\footnote{\url{https://www.robots.ox.ac.uk/∼vgg/data/flowers/}}, HandWritten\footnote{\url{https://archive.ics.uci.edu/ml/datasets/Multiple+Features}}, BDGP\footnote{\url{https://www.fruitfly.org/}}, Hdigit\footnote{\url{https://archive.ics.uci.edu/ml/datasets.php}}, AWA\footnote{https://cvml.ist.ac.at/AwA/} and YTF-10/20\footnote{\url{http://archive.ics.uci.edu/ml/datasets/YouTube+Multiview+Video+Games+Dataset}}. The detailed information of them is summarized in Table \ref{dataset}.
\begin{table}
 \begin{center}
 \caption{Datasets used in our experiments.} 
 \label{dataset}
\begin{tabular}{|c|c|c|c|}

\hline
Datasets                       & \#Samples & \#View & \#Class \\ \hline
\rowcolor[HTML]{FFFFFF} 
Flower17                      & 1360      & 7      & 17      \\ \hline
\rowcolor[HTML]{FFFFFF} 
HandWritten                   & 2000      & 2      & 10      \\ \hline
\rowcolor[HTML]{FFFFFF} 
BDGP                          & 2500      & 3      & 5       \\ \hline
\rowcolor[HTML]{FFFFFF} 
Hdigit                        & 10000     & 2      & 10      \\ \hline
\rowcolor[HTML]{FFFFFF} 
AWA                           & 30475     & 6      & 50      \\ \hline
\rowcolor[HTML]{FFFFFF} 
YTF-10 & 38654     & 4      & 10      \\ \hline
\rowcolor[HTML]{FFFFFF} 
YTF-20 & 63896     & 4      & 20      \\ \hline
\end{tabular}
\end{center}
\end{table}
\begin{table*}[]
\renewcommand\arraystretch{1.3}
\scriptsize
\centering
	\caption{Empirical evaluation and comparison of OMVCDR with seven baseline methods on seven benchmark datasets in terms of clustering accuracy (ACC), normalized mutual information (NMI), Purity, and Fscore. Note that '\--{}' indicates the algorithm cannot be executed smoothly due to the out-of-memory error or other reasons.}
	\label{result}
\begin{tabular}{|ccccccccc|}
\hline
\multicolumn{1}{|c|}{{\color[HTML]{000000} }} & \multicolumn{1}{c|}{\begin{tabular}[c]{@{}c@{}}FMR\\ \cite{ijcai2019p404}\end{tabular}} & \multicolumn{1}{c|}{\begin{tabular}[c]{@{}c@{}}SFMC\\ \cite{9146384}\end{tabular}} & \multicolumn{1}{c|}{\begin{tabular}[c]{@{}c@{}}SMVSC\\ \cite{Sun2021ScalableMS}\end{tabular}} & \multicolumn{1}{c|}{\begin{tabular}[c]{@{}c@{}}OPMC\\ \cite{9710821}\end{tabular}}& \multicolumn{1}{c|}{\begin{tabular}[c]{@{}c@{}}FPMVS\\ \cite{9646486}\end{tabular}} & \multicolumn{1}{c|}{\begin{tabular}[c]{@{}c@{}}LMVSC\\ \cite{Kang2020LargescaleMS}\end{tabular}} & \multicolumn{1}{c|}{\begin{tabular}[c]{@{}c@{}}AWMVC\\ \cite{wanAAAI2023}\end{tabular}} & \multicolumn{1}{c|}{\begin{tabular}[c]{@{}c@{}}OMVCDR \end{tabular}}                   \\ \hline
\multicolumn{9}{|c|}{\cellcolor[HTML]{FFFFFF}ACC}                                                                                                                                                                                                                                                                                                                                                                                                                    \\ \hline
\multicolumn{1}{|c|}{\cellcolor[HTML]{FFFFFF}Flower17}                      & \multicolumn{1}{c|}{\cellcolor[HTML]{FFFFFF}0.3343±0.0175} & \multicolumn{1}{c|}{\cellcolor[HTML]{FFFFFF}0.0757±0.0000} & \multicolumn{1}{c|}{\cellcolor[HTML]{FFFFFF}0.1610±0.0031} & \multicolumn{1}{c|}{0.3213±0.0000}       & \multicolumn{1}{c|}{0.2603±0.0003} & \multicolumn{1}{c|}{0.3495±0.0214}          & \multicolumn{1}{c|}{{\ul 0.3541±0.0208}} & \textbf{0.3787±0.0000} \\ \hline
\multicolumn{1}{|c|}{\cellcolor[HTML]{FFFFFF}HandWritten}                   & \multicolumn{1}{c|}{\cellcolor[HTML]{FFFFFF}0.6748±0.0459} & \multicolumn{1}{c|}{\cellcolor[HTML]{FFFFFF}0.4115±0.0000} & \multicolumn{1}{c|}{\cellcolor[HTML]{FFFFFF}0.5684±0.0356} & \multicolumn{1}{c|}{0.7670±0.0000}       & \multicolumn{1}{c|}{0.6708±0.0470} & \multicolumn{1}{c|}{{\ul 0.8440±0.0751}}    & \multicolumn{1}{c|}{0.6370±0.0252}       & \textbf{0.9250±0.0000} \\ \hline
\multicolumn{1}{|c|}{\cellcolor[HTML]{FFFFFF}BDGP}                          & \multicolumn{1}{c|}{\cellcolor[HTML]{FFFFFF}0.4184±0.0003} & \multicolumn{1}{c|}{\cellcolor[HTML]{FFFFFF}-}             & \multicolumn{1}{c|}{\cellcolor[HTML]{FFFFFF}0.3679±0.0119} & \multicolumn{1}{c|}{0.4836±0.0000}       & \multicolumn{1}{c|}{0.4242±0.0120} & \multicolumn{1}{c|}{{\ul 0.5080±0.0211}}    & \multicolumn{1}{c|}{0.3616±0.0106}       & \textbf{0.5696±0.0000} \\ \hline
\multicolumn{1}{|c|}{\cellcolor[HTML]{FFFFFF}Hdigit}                        & \multicolumn{1}{c|}{\cellcolor[HTML]{FFFFFF}-}             & \multicolumn{1}{c|}{\cellcolor[HTML]{FFFFFF}0.4773±0.0000} & \multicolumn{1}{c|}{\cellcolor[HTML]{FFFFFF}0.6494±0.0459} & \multicolumn{1}{c|}{0.8056±0.0000}       & \multicolumn{1}{c|}{0.7867±0.0519} & \multicolumn{1}{c|}{{\ul 0.8686±0.0806}}    & \multicolumn{1}{c|}{0.6570±0.0225}       & \textbf{0.9052±0.0000} \\ \hline
\multicolumn{1}{|c|}{\cellcolor[HTML]{FFFFFF}{\color[HTML]{333333} AWA}}    & \multicolumn{1}{c|}{\cellcolor[HTML]{FFFFFF}-}             & \multicolumn{1}{c|}{\cellcolor[HTML]{FFFFFF}0.0390±0.0000} & \multicolumn{1}{c|}{\cellcolor[HTML]{FFFFFF}0.0638±0.0003} & \multicolumn{1}{c|}{{\ul 0.0941±0.0000}} & \multicolumn{1}{c|}{0.0887±0.0013} & \multicolumn{1}{c|}{0.0768±0.0016}          & \multicolumn{1}{c|}{0.0864±0.0022}       & \textbf{0.0955±0.0000} \\ \hline
\multicolumn{1}{|c|}{\cellcolor[HTML]{FFFFFF}{\color[HTML]{333333} YTF-10}} & \multicolumn{1}{c|}{\cellcolor[HTML]{FFFFFF}-}             & \multicolumn{1}{c|}{\cellcolor[HTML]{FFFFFF}0.5580±0.0000} & \multicolumn{1}{c|}{\cellcolor[HTML]{FFFFFF}0.6300±0.0434} & \multicolumn{1}{c|}{{\ul 0.7613±0.0000}} & \multicolumn{1}{c|}{0.6741±0.0359} & \multicolumn{1}{c|}{0.7389±0.0543}          & \multicolumn{1}{c|}{0.6738±0.0687}       & \textbf{0.7882±0.0000} \\ \hline
\multicolumn{1}{|c|}{YTF-20}                                                & \multicolumn{1}{c|}{-}                                     & \multicolumn{1}{c|}{-}                                     & \multicolumn{1}{c|}{0.5677±0.0466}                         & \multicolumn{1}{c|}{0.6544±0.0000}       & \multicolumn{1}{c|}{0.6259±0.0271} & \multicolumn{1}{c|}{{\ul 0.6834±0.0422}}    & \multicolumn{1}{c|}{0.6612±0.0478}       & \textbf{0.7418±0.0000} \\ \hline
\multicolumn{9}{|c|}{NMI}                                                                                                                                                                                                                                                                                                                                                                                                                                            \\ \hline
\multicolumn{1}{|c|}{\cellcolor[HTML]{FFFFFF}Flower17}                      & \multicolumn{1}{c|}{0.3065±0.0091}                         & \multicolumn{1}{c|}{0.0787±0.0000}                         & \multicolumn{1}{c|}{0.1219±0.0038}                         & \multicolumn{1}{c|}{0.2969±0.0000}       & \multicolumn{1}{c|}{0.2603±0.0009} & \multicolumn{1}{c|}{0.3404±0.0156}          & \multicolumn{1}{c|}{{\ul 0.3425±0.0152}} & \textbf{0.4915±0.0000} \\ \hline
\multicolumn{1}{|c|}{\cellcolor[HTML]{FFFFFF}HandWritten}                   & \multicolumn{1}{c|}{0.5900±0.0178}                         & \multicolumn{1}{c|}{0.4500±0.0000}                         & \multicolumn{1}{c|}{0.5239±0.0518}                         & \multicolumn{1}{c|}{0.7016±0.0000}       & \multicolumn{1}{c|}{0.6592±0.0243} & \multicolumn{1}{c|}{\textbf{0.8524±0.0256}} & \multicolumn{1}{c|}{0.5926±0.0195}       & {\ul 0.8507±0.0000}    \\ \hline
\multicolumn{1}{|c|}{\cellcolor[HTML]{FFFFFF}BDGP}                          & \multicolumn{1}{c|}{0.1257±0.0005}                         & \multicolumn{1}{c|}{-}                                     & \multicolumn{1}{c|}{{\ul 0.3404±0.0187}}                   & \multicolumn{1}{c|}{0.2591±0.0000}       & \multicolumn{1}{c|}{0.2102±0.0166} & \multicolumn{1}{c|}{0.2604±0.0175}          & \multicolumn{1}{c|}{0.1093±0.0099}       & \textbf{0.3558±0.0000} \\ \hline
\multicolumn{1}{|c|}{\cellcolor[HTML]{FFFFFF}Hdigit}                        & \multicolumn{1}{c|}{-}                                     & \multicolumn{1}{c|}{0.6362±0.0000}                         & \multicolumn{1}{c|}{0.5884±0.0690}                         & \multicolumn{1}{c|}{0.7480±0.0000}       & \multicolumn{1}{c|}{0.7700±0.0214} & \multicolumn{1}{c|}{\textbf{0.8933±0.0306}} & \multicolumn{1}{c|}{0.6079±0.0151}       & {\ul 0.8755±0.0000}    \\ \hline
\multicolumn{1}{|c|}{\cellcolor[HTML]{FFFFFF}{\color[HTML]{333333} AWA}}    & \multicolumn{1}{c|}{-}                                     & \multicolumn{1}{c|}{0.0032±0.0000}                         & \multicolumn{1}{c|}{0.0415±0.0005}                         & \multicolumn{1}{c|}{{\ul 0.1195±0.0000}} & \multicolumn{1}{c|}{0.1037±0.0025} & \multicolumn{1}{c|}{0.0897±0.0017}          & \multicolumn{1}{c|}{0.1049±0.0024}       & \textbf{0.1472±0.0000} \\ \hline
\multicolumn{1}{|c|}{\cellcolor[HTML]{FFFFFF}{\color[HTML]{333333} YTF-10}} & \multicolumn{1}{c|}{-}                                     & \multicolumn{1}{c|}{0.7746±0.0000}                         & \multicolumn{1}{c|}{0.5438±0.0637}                         & \multicolumn{1}{c|}{{\ul 0.8003±0.0000}} & \multicolumn{1}{c|}{0.7343±0.0242} & \multicolumn{1}{c|}{0.7666±0.0224}          & \multicolumn{1}{c|}{0.7638±0.0368}       & \textbf{0.8072±0.0000} \\ \hline
\multicolumn{1}{|c|}{YTF-20}                                                & \multicolumn{1}{c|}{-}                                     & \multicolumn{1}{c|}{-}                                     & \multicolumn{1}{c|}{0.4899±0.0679}                         & \multicolumn{1}{c|}{0.7725±0.0000}       & \multicolumn{1}{c|}{0.7496±0.0173} & \multicolumn{1}{c|}{0.7709±0.0153}          & \multicolumn{1}{c|}{{\ul 0.7728±0.0180}} & \textbf{0.7831±0.0000} \\ \hline
\multicolumn{9}{|c|}{Purity}                                                                                                                                                                                                                                                                                                                                                                                                                                         \\ \hline
\multicolumn{1}{|c|}{\cellcolor[HTML]{FFFFFF}Flower17}                      & \multicolumn{1}{c|}{0.3474±0.0138}                         & \multicolumn{1}{c|}{0.1029±0.0000}                         & \multicolumn{1}{c|}{0.2372±0.0061}                         & \multicolumn{1}{c|}{0.3360±0.0000}       & \multicolumn{1}{c|}{0.2742±0.0005} & \multicolumn{1}{c|}{0.3618±0.0212}          & \multicolumn{1}{c|}{{\ul 0.3677±0.0205}} & \textbf{0.4125±0.0000} \\ \hline
\multicolumn{1}{|c|}{\cellcolor[HTML]{FFFFFF}HandWritten}                   & \multicolumn{1}{c|}{0.6869±0.0316}                         & \multicolumn{1}{c|}{0.4170±0.0000}                         & \multicolumn{1}{c|}{0.6239±0.0128}                         & \multicolumn{1}{c|}{0.7670±0.0000}       & \multicolumn{1}{c|}{0.6768±0.0465} & \multicolumn{1}{c|}{{\ul 0.8654±0.0591}}    & \multicolumn{1}{c|}{0.6590±0.0243}       & \textbf{0.9250±0.0000} \\ \hline
\multicolumn{1}{|c|}{\cellcolor[HTML]{FFFFFF}BDGP}                          & \multicolumn{1}{c|}{0.4236±0.0006}                         & \multicolumn{1}{c|}{-}                                     & \multicolumn{1}{c|}{0.4021±0.02332}                        & \multicolumn{1}{c|}{0.4900±0.0000}       & \multicolumn{1}{c|}{0.4292±0.0124} & \multicolumn{1}{c|}{{\ul 0.5088±0.0169}}    & \multicolumn{1}{c|}{0.3762±0.0053}       & \textbf{0.5936±0.0000} \\ \hline
\multicolumn{1}{|c|}{\cellcolor[HTML]{FFFFFF}Hdigit}                        & \multicolumn{1}{c|}{-}                                     & \multicolumn{1}{c|}{0.4781±0.0000}                         & \multicolumn{1}{c|}{0.7295±0.0143}                         & \multicolumn{1}{c|}{0.8058±0.0000}       & \multicolumn{1}{c|}{0.7872±0.0518} & \multicolumn{1}{c|}{{\ul 0.8893±0.0635}}    & \multicolumn{1}{c|}{0.6734±0.0220}       & \textbf{0.9052±0.0000} \\ \hline
\multicolumn{1}{|c|}{\cellcolor[HTML]{FFFFFF}{\color[HTML]{333333} AWA}}    & \multicolumn{1}{c|}{-}                                     & \multicolumn{1}{c|}{0.0399±0.0000}                         & \multicolumn{1}{c|}{\textbf{0.1386±0.0023}}                & \multicolumn{1}{c|}{{\ul 0.1140±0.0000}} & \multicolumn{1}{c|}{0.0926±0.0015} & \multicolumn{1}{c|}{0.0986±0.0019}          & \multicolumn{1}{c|}{0.1052±0.0027}       & 0.1105±0.0000          \\ \hline
\multicolumn{1}{|c|}{\cellcolor[HTML]{FFFFFF}{\color[HTML]{333333} YTF-10}} & \multicolumn{1}{c|}{-}                                     & \multicolumn{1}{c|}{0.7410±0.0000}                         & \multicolumn{1}{c|}{0.7551±0.0185}                         & \multicolumn{1}{c|}{{\ul 0.7842±0.0000}} & \multicolumn{1}{c|}{0.7027±0.0389} & \multicolumn{1}{c|}{0.7674±0.0336}          & \multicolumn{1}{c|}{0.7542±0.0499}       & \textbf{0.8156±0.0000} \\ \hline
\multicolumn{1}{|c|}{YTF-20}                                                & \multicolumn{1}{c|}{-}                                     & \multicolumn{1}{c|}{-}                                     & \multicolumn{1}{c|}{0.6828±0.0121}                         & \multicolumn{1}{c|}{{\ul 0.7475±0.0000}} & \multicolumn{1}{c|}{0.6633±0.0330} & \multicolumn{1}{c|}{0.7389±0.0287}          & \multicolumn{1}{c|}{0.7285±0.0331}       & \textbf{0.7771±0.0000} \\ \hline
\multicolumn{9}{|c|}{Fscore}                                                                                                                                                                                                                                                                                                                                                                                                                                         \\ \hline
\multicolumn{1}{|c|}{\cellcolor[HTML]{FFFFFF}Flower17}                      & \multicolumn{1}{c|}{0.2009±0.0081}                         & \multicolumn{1}{c|}{0.1094±0.0000}                         & \multicolumn{1}{c|}{{\ul 0.2403±0.0102}}                   & \multicolumn{1}{c|}{0.1826±0.0000}       & \multicolumn{1}{c|}{0.1665±0.0009} & \multicolumn{1}{c|}{0.2274±0.0127}          & \multicolumn{1}{c|}{0.2336±0.0139}       & \textbf{0.3046±0.0000} \\ \hline
\multicolumn{1}{|c|}{\cellcolor[HTML]{FFFFFF}HandWritten}                   & \multicolumn{1}{c|}{0.5534±0.0316}                         & \multicolumn{1}{c|}{0.2968±0.0000}                         & \multicolumn{1}{c|}{0.6121±0.0298}                         & \multicolumn{1}{c|}{0.6654±0.0000}       & \multicolumn{1}{c|}{0.5944±0.0364} & \multicolumn{1}{c|}{{\ul 0.8171±0.0565}}    & \multicolumn{1}{c|}{0.6590±0.0229}       & \textbf{0.8561±0.0000} \\ \hline
\multicolumn{1}{|c|}{\cellcolor[HTML]{FFFFFF}BDGP}                          & \multicolumn{1}{c|}{0.2921±0.0004}                         & \multicolumn{1}{c|}{-}                                     & \multicolumn{1}{c|}{0.2375±0.0251}                         & \multicolumn{1}{c|}{0.3787±0.0000}       & \multicolumn{1}{c|}{0.3455±0.0044} & \multicolumn{1}{c|}{{\ul 0.3907±0.0095}}    & \multicolumn{1}{c|}{0.2771±0.0057}       & \textbf{0.4610±0.0000} \\ \hline
\multicolumn{1}{|c|}{\cellcolor[HTML]{FFFFFF}Hdigit}                        & \multicolumn{1}{c|}{-}                                     & \multicolumn{1}{c|}{0.4774±0.0000}                         & \multicolumn{1}{c|}{0.6943±0.0304}                         & \multicolumn{1}{c|}{0.7386±0.0000}       & \multicolumn{1}{c|}{0.7340±0.0441} & \multicolumn{1}{c|}{\textbf{0.8562±0.0721}} & \multicolumn{1}{c|}{0.5561±0.0181}       & {\ul 0.8241±0.0000}    \\ \hline
\multicolumn{1}{|c|}{\cellcolor[HTML]{FFFFFF}{\color[HTML]{333333} AWA}}    & \multicolumn{1}{c|}{-}                                     & \multicolumn{1}{c|}{0.0457±0.0000}                         & \multicolumn{1}{c|}{\textbf{0.1045±0.0022}}                & \multicolumn{1}{c|}{0.0469±0.0000}       & \multicolumn{1}{c|}{0.0632±0.0009} & \multicolumn{1}{c|}{0.0386±0.0006}          & \multicolumn{1}{c|}{0.0423±0.0009}       & {\ul 0.0668±0.0000}    \\ \hline
\multicolumn{1}{|c|}{\cellcolor[HTML]{FFFFFF}{\color[HTML]{333333} YTF-10}} & \multicolumn{1}{c|}{-}                                     & \multicolumn{1}{c|}{0.6125±0.0000}                         & \textbf {0.7478±0.0300}                   & \multicolumn{1}{c|}{0.7217±0.0000}       & \multicolumn{1}{c|}{0.6305±0.0422} & \multicolumn{1}{c|}{0.6671±0.0346}          & \multicolumn{1}{c|}{0.6563±0.0691}       &  \multicolumn{1}{c|}{{\ul 0.7436±0.0000}} \\ \hline
\multicolumn{1}{|c|}{YTF-20}                                                & \multicolumn{1}{c|}{-}                                     & \multicolumn{1}{c|}{-}                                     & \multicolumn{1}{c|}{{\ul 0.6383±0.0238}}                   & \multicolumn{1}{c|}{0.6073±0.0000}       & \multicolumn{1}{c|}{0.5715±0.0366} & \multicolumn{1}{c|}{0.6378±0.0407}          & \multicolumn{1}{c|}{0.6149±0.0452}       & \textbf{0.6391±0.0000} \\ \hline
\end{tabular}
\end{table*}

\subsubsection{Compared Algorithms}
\begin{enumerate}
 \item \textbf{Flexible Multi-View Representation Learning for Subspace Clustering (FMR)} \cite{ijcai2019p404}. It conducts subspace clustering and attains comprehensive information by enforcing the latent representation among views close to each other.
 
\item \textbf{Multiview Clustering: A Scalable and Parameter-Free Bipartite Graph Fusion Method (SFMC)}\cite{9146384}. SFMC seeks a joint graph compatible across multiple views in a self-supervised weighting manner.

 \item \textbf{Scalable Multi-view Subspace Clustering with Unified Anchors (SMVSC)} \cite{Sun2021ScalableMS}. It conducts anchor learning among views with unified anchors and puts anchor learning and subspace learning into a unified optimization framework.
 
\item \textbf{One-pass Multi-view Clustering for Large-scale Data (OPMC)} \cite{9710821}. OPMC removes the non-negativity constraint of MVCNMF and proposes a one-pass multi-view clustering algorithm.

\item \textbf{Fast Parameter-Free Multi-View Subspace Clustering With Consensus Anchor Guidance (FPMVS)} \cite{9646486}. It simultaneously conducts anchor selection and subspace graph construction in a parameter-free manner.

\item \textbf{Large-scale Multi-view Subspace Clustering in Linear Time (LMVSC)} \cite{Kang2020LargescaleMS}. LMVSC integrates anchor graphs and conducts spectral clustering on a smaller graph.

\item \textbf{Auto-weighted Multi-view Clustering for Large-scale Data (AWMVC)} \cite{wanAAAI2023}. AWMVC learns coefficient matrices from corresponding base matrices of different dimensions to enhance the representation ability.                                           
 \end{enumerate}
 \begin{figure*}[htbp]
	\centering
	\subfigure{
		\includegraphics[width=0.8\textwidth]{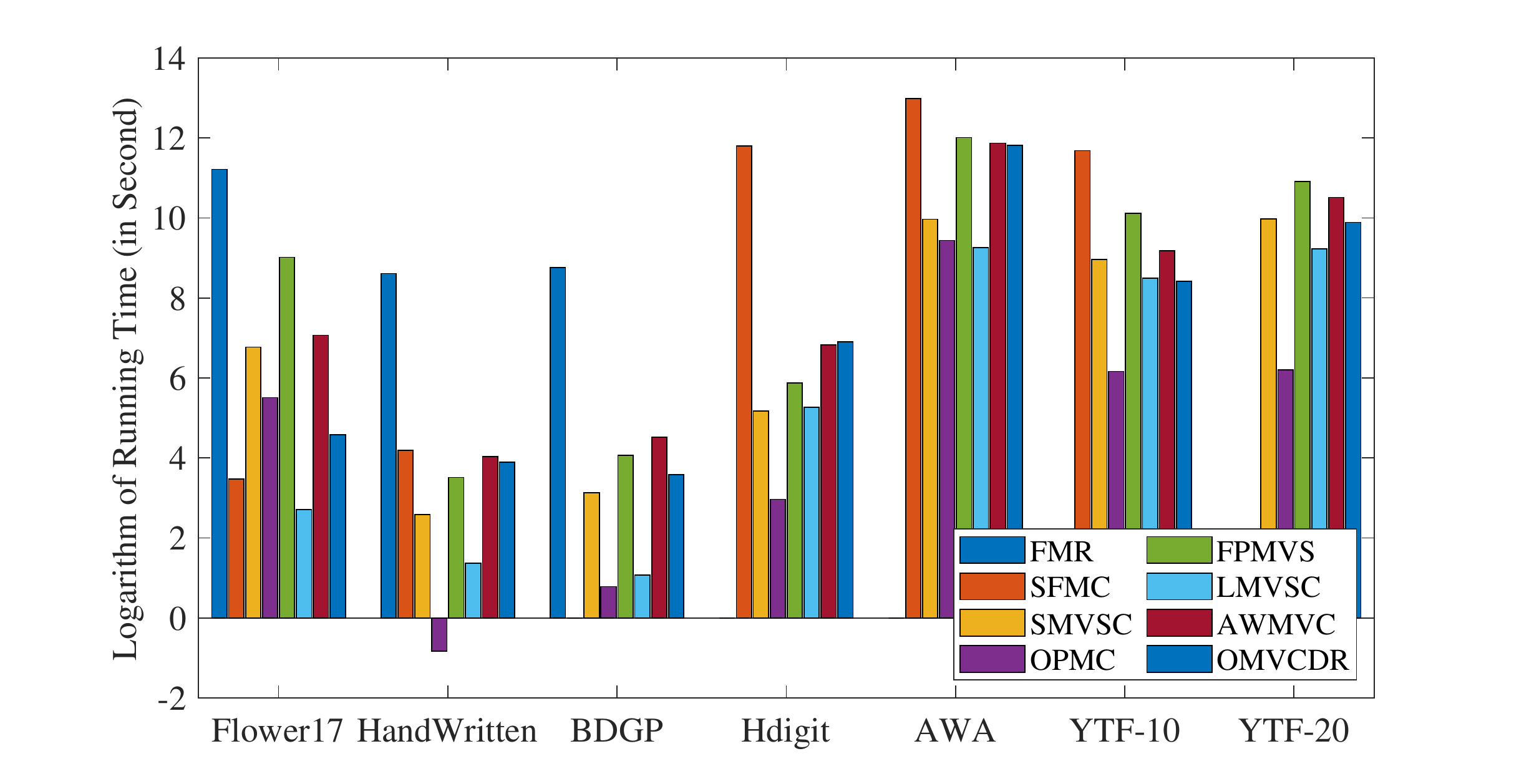}}

	\caption{The relative running time of the compared algorithms on the seven benchmark datasets. The empty bar indicates that the respective method fails to run smoothly due to an out-of-memory error or other reasons.}
	\label{running_time}
\end{figure*}
\subsubsection{Metrics}
Four popular clustering indicators are employed to evaluate the clustering performance of our proposed method, including accuracy (ACC), normalized mutual information (NMI), purity and Fscore. These metrics measure the consistency between true labels and predicted labels from different aspects and higher values frequently indicate better clustering quality.

For given $\mathbf{x}_i \left(1\leq i\leq n\right)$, let $c_i$ and $y_i$ denote the predicted label and ground-truth label of $i$-th sample, separately. ACC can be obtained as follows,
\begin{equation}
\begin{aligned}
\operatorname{ACC}=\frac{\sum_{i=1}^n \delta\left(y_i, \operatorname{map}\left(c_i\right)\right)}{n},
\end{aligned}
\end{equation}
where $\delta\left(u,v\right)$ is the delta function that outputs one if $u=v$ and outputs zero otherwise, $\operatorname{map}\left(c_i\right)$ is the permutation mapping function that maps $c_i$ to the true label according to Kuhn-Munkres algorithm \cite{Kuhn1955TheHM}.

The acquisition of NMI is based on the mutual information (MI) between $\mathbf{y}$ and $\mathbf{c}$, which is defined as follows,
\begin{equation}
\begin{aligned}
\operatorname{MI}(\mathbf{y}, \mathbf{c})=\sum_{y_i \in \mathbf{y}, c_j^{\prime} \in \mathbf{c}} p\left(y_i, c_j^{\prime}\right) \log _2 \frac{p\left(y_i, c_j^{\prime}\right)}{p\left(y_i\right) p\left(c_j^{\prime}\right)},
\end{aligned}
\end{equation}
where $p\left(y_i\right)$ and $p\left( c_j^{\prime}\right)$ are the probabilities that a sample is randomly selected from cluster $y_i$ and $c_j$, respectively, and $p\left(y_i, c_j^{\prime}\right)$ is the joint probability that a sample is arbitrarily chosen from clusters $y_i$ and $c_j$, simultaneously. Based on this, NMI is computed by
\begin{equation}
\begin{aligned}
\operatorname{NMI}(\mathbf{y}, \mathbf{c})=\frac{\operatorname{MI}(\mathbf{y}, \mathbf{c})}{\max (\mathrm{H}(\mathbf{y}), \mathrm{H}(\mathbf{c}))},
\end{aligned}
\end{equation}
where $\mathrm{H}(\mathbf{y})$ and $\mathrm{H}(\mathbf{c})$ are the  entropies of $\mathbf{y}$ and $\mathbf{c}$, individually.

Let $\mathcal{S}$ and $\mathcal{S}^{\prime}$ denote the real label set and the predicted label set, separately, purity is defined as follows,
\begin{equation}
\begin{aligned}
\operatorname{Purity}\left(\mathcal{S}, \mathcal{S}^{\prime}\right)=\frac{1}{n} \sum_k \max _j\left|\mathcal{S}_j \cap \mathcal{S}_k^{\prime}\right|.
\end{aligned}
\end{equation}

As for Fscore, it it obtained by
\begin{equation}
\begin{aligned}
\operatorname{Fscore}=\frac{2*\operatorname{precision}*\operatorname{recall}}{\operatorname {precision}+\operatorname{recall}}.
\end{aligned}
\end{equation}

\subsubsection{Schemes}
The codes of the compared algorithms are publicly available online, and we run them without any changes. Given that some methods separate representation learning and clustering into two processes, we perform k-means 50 times and report the averages to eliminate the randomness in initialization. In our experiment, we simply s
`et $m=3$ and the dimensions range from $k$ to $mk$. We tune the hyper-parameter $\lambda \in 2 .^{\wedge}[-5,-4, \cdots, 5]$. All the experiments are conducted on a desktop computer with Intel(R) Core(TM) i9-10850K CPU and 96G RAM.
\subsubsection{Initialization}
In our implementation, $\mathbf{H}_p^{(v)}$ is generated by setting the first $d_p$ diagonal elements to 1 and other elements to 0. We initialize $\mathbf{Y}$ by performing single-view clustering on $\left[\mathbf{X}^{(1)};\dots;\mathbf{X}^{(V)}\right]$, where $\left[\cdot;\cdot\right]$ is the vertical concatenation operation. As for the production of  $\boldsymbol{\alpha}$, we set each element with an equal value. 

\subsection{Experimental Results}

We compare our proposed method with seven algorithms on seven widely used benchmark datasets according to four clustering metrics, and the results are demonstrated in Table \ref{result}. It is worth noting that the best outcome is marked in bold, and the second best is underlined. In addition, '\--{}' indicates that the algorithm fails to run smoothly due to an out-of-memory error or other reasons. From the table, we have the following observations:                    
\begin{enumerate}
\item Our proposed OMVCDR demonstrates its superiority over the competitors. Especially in terms of ACC, it consistently outperforms the second best algorithm over 6.95\%, 9.60\%, 12.13\%, 4.21\%, 1.49\%, 3.53\% and 8.55\% on all datasets, respectively. Furthermore,  it shows comparable performance on other metrics as well. The promising outcome supports the effectiveness of OMVCDR.
\item  Compared with FMR, SMVSC, FPMVS, LMVSC, and AWMVC, which undergo a two-step process, i.e., representation learning and the subsequent $k$-means, the variance of OMVCDR is zero. Given that we propose a one-pass algorithm and the clustering label is directly  generated, the clustering results are stable. This property displays the robustness of OMVCDR.
\item In comparison to SFMC and OPMC, which are under a one-pass framework, OMVCDR also performs better. Owing to mapping original features into diverse dimensions and automatically choosing a suitable latent space, our proposed method is accomplished in representation learning. On the contrary, SFMC gets a dimension-fixed anchor graph and fails to connect view-specific anchor learning with consensus graph learning, and OPMC gets trouble selecting a proper space for each view.
\item Most large-scale algorithms fail to attain good clustering results on regular datasets, and OMVCDR overcomes this drawback. For instance, anchor-based methods always utilize anchor points to represent the data cloud. However, the number of anchors gets scarce when handling regular datasets and cannot ensure enough information is extracted. Thus, their performance declines in these scenarios. Conversely, OMVCDR is suitable for diverse applications since it attains more comprehensive knowledge.
\end{enumerate}
 \begin{figure*}[htbp]
	\centering

	\subfigure{
		\includegraphics[width=0.23\textwidth]{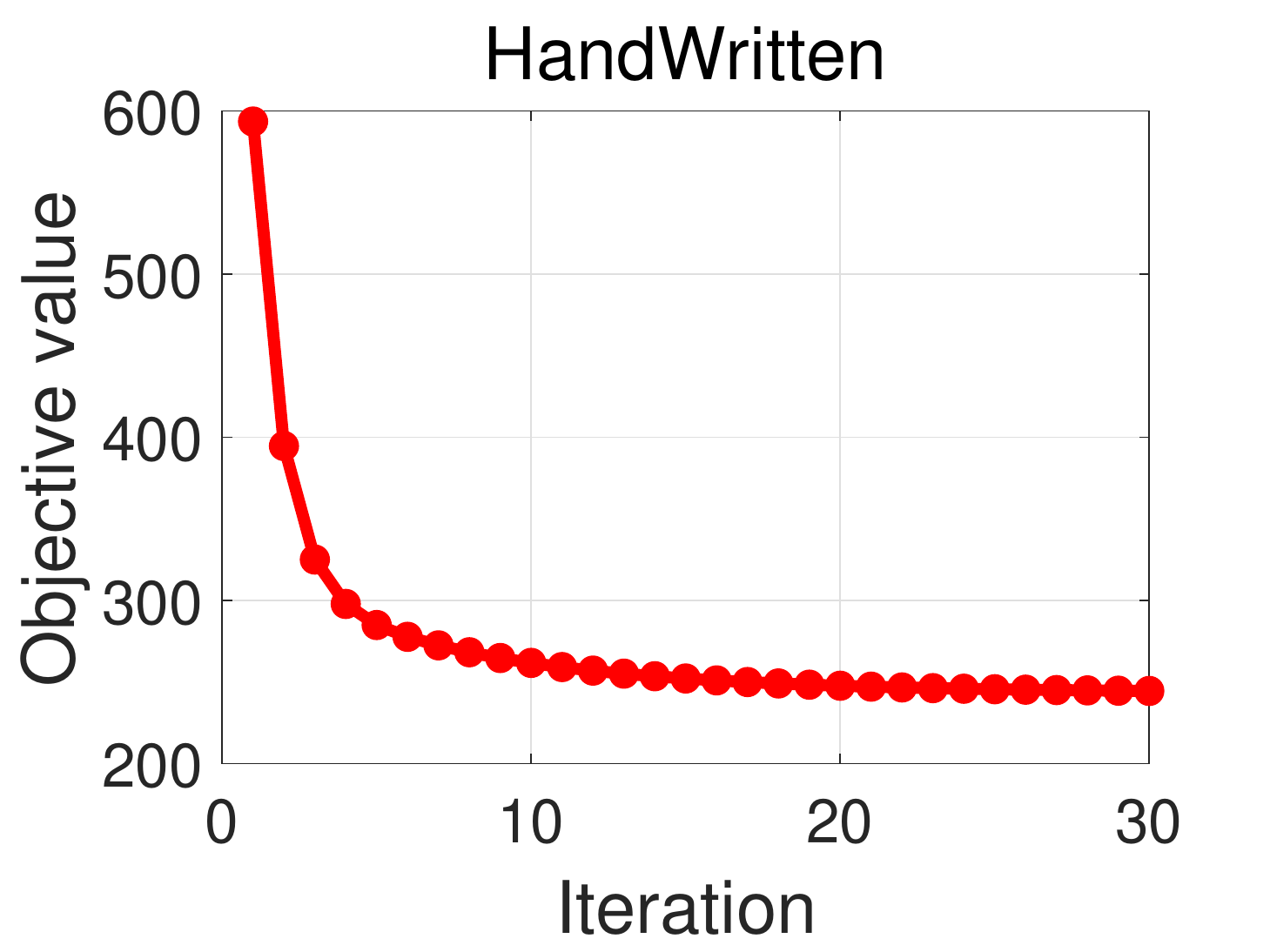}}
	\subfigure{
		\includegraphics[width=0.23\textwidth]{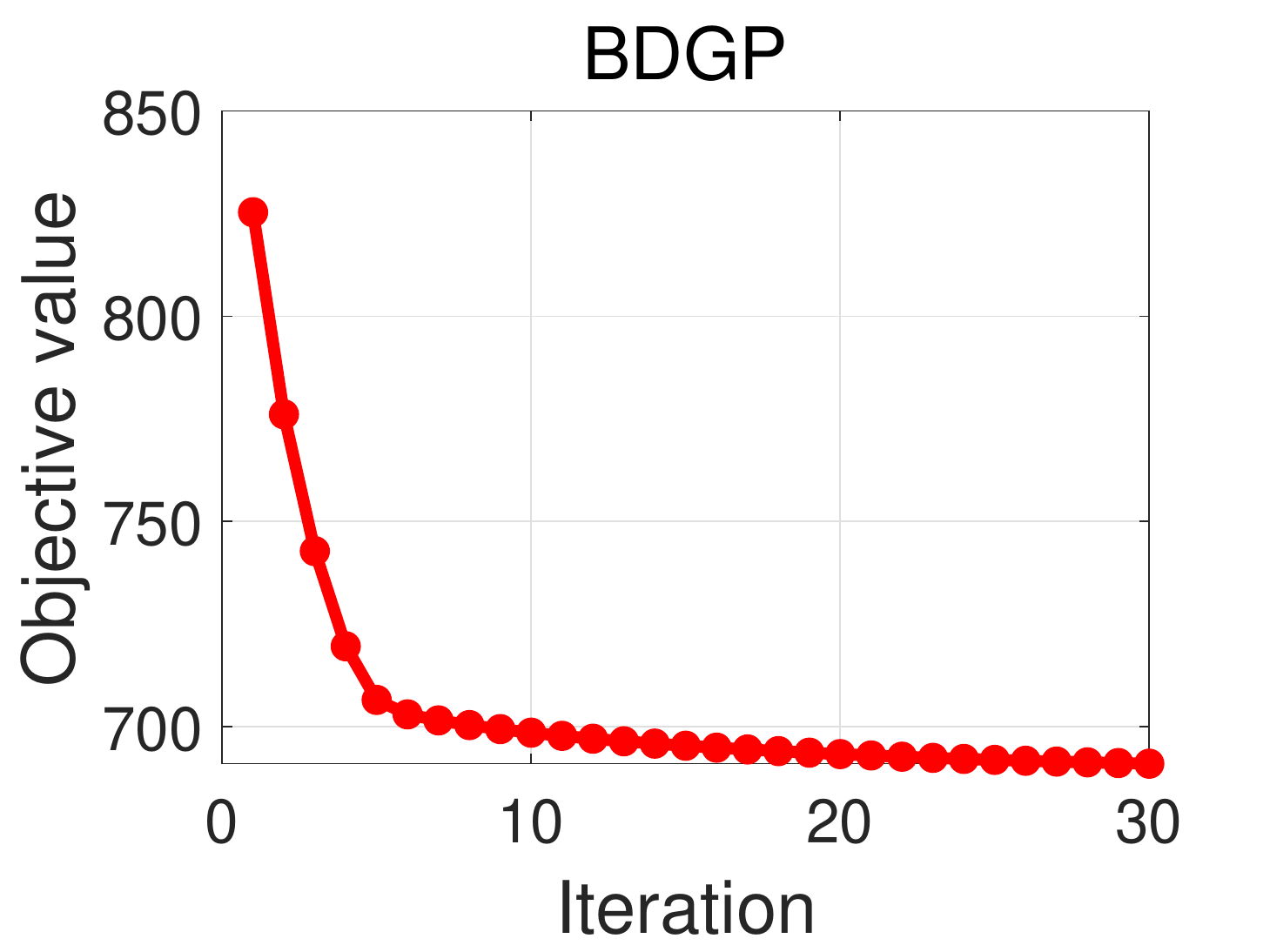}}
	\subfigure{
		\includegraphics[width=0.23\textwidth]{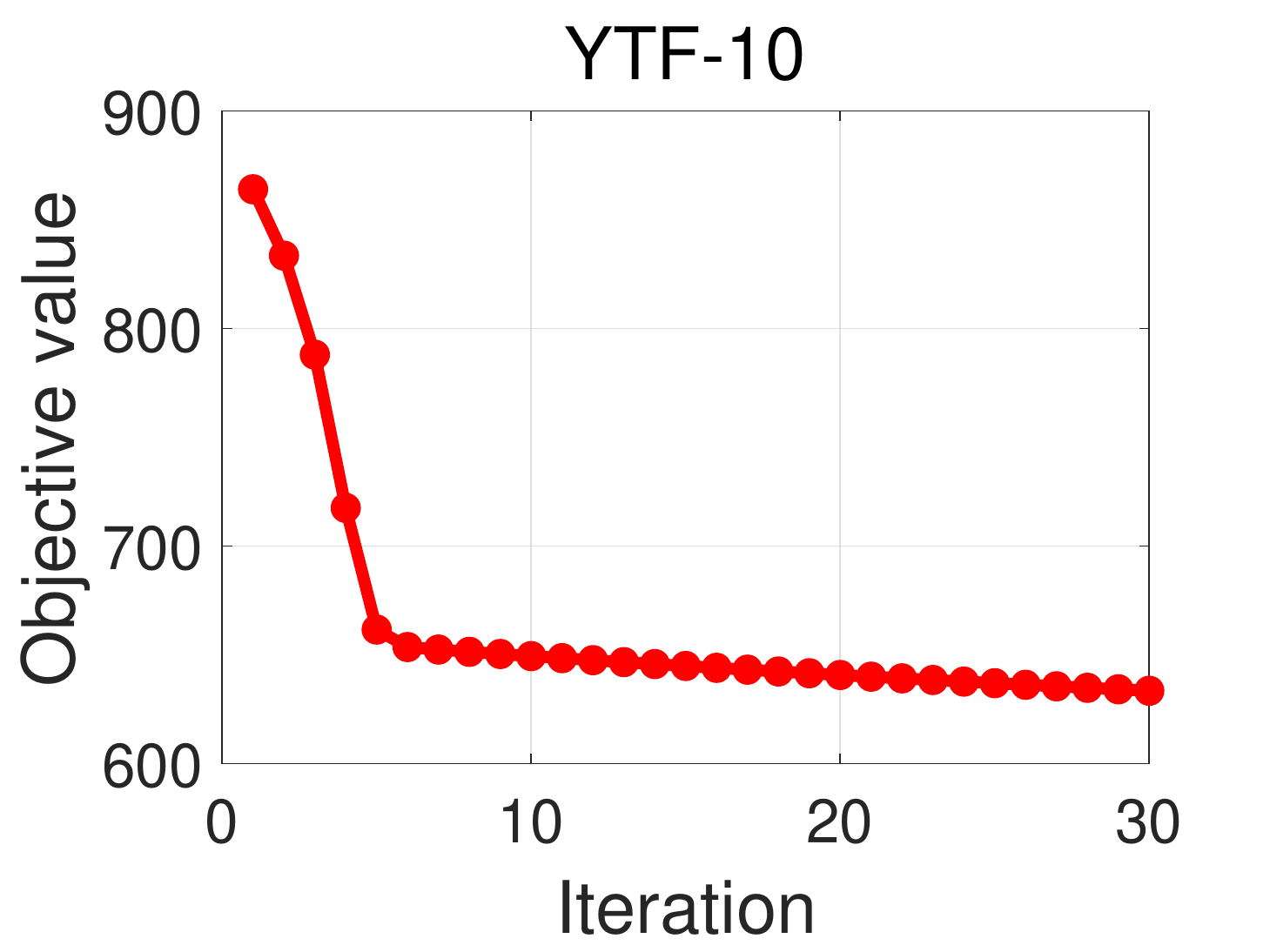}}
	\subfigure{
		\includegraphics[width=0.23\textwidth]{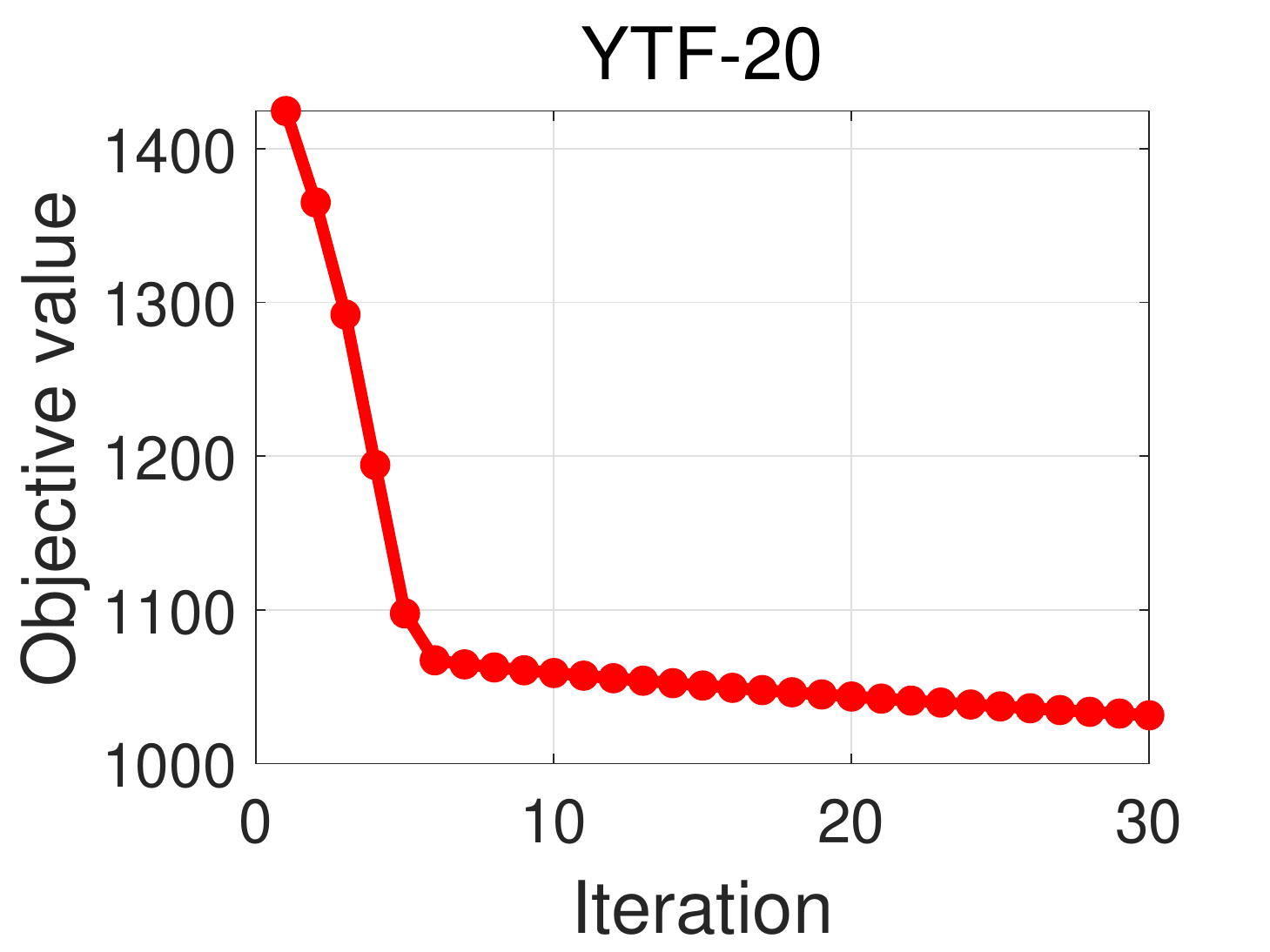}}

	
\centering
	\caption{The objective values of OMVCDR vary with iterations on four benchmark datasets, including HandWritten, BDGP, YTF-10, and YTF-20, respectively.}
\label{fig_obj}
\end{figure*}

\begin{table}[]
\renewcommand\arraystretch{1.2}
\centering
\caption{The ablation study of our proposed method on seven benchmark datasets in terms of ACC, NMI, Purity and Fscore. The best results are marked in bold.}\label{ablation_study}

\begin{tabular}{|ccccc|}
\hline
\multicolumn{1}{|c|}{Datasets}    & \multicolumn{1}{c|}{OMVC}    & \multicolumn{1}{c|}{OMVCDR-2}    & \multicolumn{1}{c|}{OMVCDR-$\boldsymbol{\alpha}$}             & Ours            \\ \hline
\multicolumn{5}{|c|}{ACC}                                                                                                                              \\ \hline
\multicolumn{1}{|c|}{Flower17}    & \multicolumn{1}{c|}{0.3537} & \multicolumn{1}{c|}{0.3582} & \multicolumn{1}{c|}{0.3610}          & \textbf{0.3787} \\ \hline
\multicolumn{1}{|c|}{HandWritten} & \multicolumn{1}{c|}{0.4715} & \multicolumn{1}{c|}{0.9245} & \multicolumn{1}{c|}{0.6720}          & \textbf{0.9250} \\ \hline
\multicolumn{1}{|c|}{BDGP}        & \multicolumn{1}{c|}{0.4020} & \multicolumn{1}{c|}{0.5564} & \multicolumn{1}{c|}{0.4820}          & \textbf{0.5696} \\ \hline
\multicolumn{1}{|c|}{Hdigit}      & \multicolumn{1}{c|}{0.3559} & \multicolumn{1}{c|}{0.8299} & \multicolumn{1}{c|}{0.5605}          & \textbf{0.9052} \\ \hline
\multicolumn{1}{|c|}{AWA}         & \multicolumn{1}{c|}{0.0934} & \multicolumn{1}{c|}{0.0958} & \multicolumn{1}{c|}{\textbf{0.0979}} & 0.0955          \\ \hline
\multicolumn{1}{|c|}{YTF-10}      & \multicolumn{1}{c|}{0.5725} & \multicolumn{1}{c|}{0.7306} & \multicolumn{1}{c|}{0.5820}          & \textbf{0.7882} \\ \hline
\multicolumn{1}{|c|}{YTF-20}      & \multicolumn{1}{c|}{0.4439} & \multicolumn{1}{c|}{0.6578} & \multicolumn{1}{c|}{0.4945}          & \textbf{0.7418} \\ \hline
\multicolumn{5}{|c|}{NMI}                                                                                                                              \\ \hline
\multicolumn{1}{|c|}{Flower17}    & \multicolumn{1}{c|}{0.4339} & \multicolumn{1}{c|}{0.4320} & \multicolumn{1}{c|}{0.4869}          & \textbf{0.4915} \\ \hline
\multicolumn{1}{|c|}{HandWritten} & \multicolumn{1}{c|}{0.4626} & \multicolumn{1}{c|}{0.8415} & \multicolumn{1}{c|}{0.6180}          & \textbf{0.8507} \\ \hline
\multicolumn{1}{|c|}{BDGP}        & \multicolumn{1}{c|}{0.1715} & \multicolumn{1}{c|}{0.3299} & \multicolumn{1}{c|}{0.2542}          & \textbf{0.3558} \\ \hline
\multicolumn{1}{|c|}{Hdigit}      & \multicolumn{1}{c|}{0.3228} & \multicolumn{1}{c|}{0.7531} & \multicolumn{1}{c|}{0.5410}          & \textbf{0.8755} \\ \hline
\multicolumn{1}{|c|}{AWA}         & \multicolumn{1}{c|}{0.1507} & \multicolumn{1}{c|}{0.1452} & \multicolumn{1}{c|}{\textbf{0.1618}} & 0.1472          \\ \hline
\multicolumn{1}{|c|}{YTF-10}      & \multicolumn{1}{c|}{0.6176} & \multicolumn{1}{c|}{0.7618} & \multicolumn{1}{c|}{0.6195}          & \textbf{0.8072} \\ \hline
\multicolumn{1}{|c|}{YTF-20}      & \multicolumn{1}{c|}{0.6214} & \multicolumn{1}{c|}{0.7454} & \multicolumn{1}{c|}{0.6733}          & \textbf{0.7831} \\ \hline
\multicolumn{5}{|c|}{Purity}                                                                                                                           \\ \hline
\multicolumn{1}{|c|}{Flower17}    & \multicolumn{1}{c|}{0.3603} & \multicolumn{1}{c|}{0.3625} & \multicolumn{1}{c|}{0.3794}          & \textbf{0.4125} \\ \hline
\multicolumn{1}{|c|}{HandWritten} & \multicolumn{1}{c|}{0.5155} & \multicolumn{1}{c|}{0.9245} & \multicolumn{1}{c|}{0.6860}          & \textbf{0.9250} \\ \hline
\multicolumn{1}{|c|}{BDGP}        & \multicolumn{1}{c|}{0.4224} & \multicolumn{1}{c|}{0.5764} & \multicolumn{1}{c|}{0.5020}          & \textbf{0.5936} \\ \hline
\multicolumn{1}{|c|}{Hdigit}      & \multicolumn{1}{c|}{0.4059} & \multicolumn{1}{c|}{0.8374} & \multicolumn{1}{c|}{0.5876}          & \textbf{0.9052} \\ \hline
\multicolumn{1}{|c|}{AWA}         & \multicolumn{1}{c|}{0.1088} & \multicolumn{1}{c|}{0.1090} & \multicolumn{1}{c|}{0.1104}          & \textbf{0.1105} \\ \hline
\multicolumn{1}{|c|}{YTF-10}      & \multicolumn{1}{c|}{0.6270} & \multicolumn{1}{c|}{0.7747} & \multicolumn{1}{c|}{0.6317}          & \textbf{0.8156} \\ \hline
\multicolumn{1}{|c|}{YTF-20}      & \multicolumn{1}{c|}{0.5455} & \multicolumn{1}{c|}{0.7138} & \multicolumn{1}{c|}{0.6272}          & \textbf{0.7771} \\ \hline
\multicolumn{5}{|c|}{Fscore}                                                                                                                           \\ \hline
\multicolumn{1}{|c|}{Flower17}    & \multicolumn{1}{c|}{0.3043} & \multicolumn{1}{c|}{0.2813} & \multicolumn{1}{c|}{0.3017}          & \textbf{0.3046} \\ \hline
\multicolumn{1}{|c|}{HandWritten} & \multicolumn{1}{c|}{0.3814} & \multicolumn{1}{c|}{0.8466} & \multicolumn{1}{c|}{0.5705}          & \textbf{0.8561} \\ \hline
\multicolumn{1}{|c|}{BDGP}        & \multicolumn{1}{c|}{0.3122} & \multicolumn{1}{c|}{0.4318} & \multicolumn{1}{c|}{0.3836}          & \textbf{0.4610} \\ \hline
\multicolumn{1}{|c|}{Hdigit}      & \multicolumn{1}{c|}{0.2676} & \multicolumn{1}{c|}{0.7510} & \multicolumn{1}{c|}{0.4734}          & \textbf{0.8241} \\ \hline
\multicolumn{1}{|c|}{AWA}         & \multicolumn{1}{c|}{0.0733} & \multicolumn{1}{c|}{0.0704} & \multicolumn{1}{c|}{\textbf{0.0756}} & 0.0668          \\ \hline
\multicolumn{1}{|c|}{YTF-10}      & \multicolumn{1}{c|}{0.5035} & \multicolumn{1}{c|}{0.7047} & \multicolumn{1}{c|}{0.5245}          & \textbf{0.7436} \\ \hline
\multicolumn{1}{|c|}{YTF-20}      & \multicolumn{1}{c|}{0.4048} & \multicolumn{1}{c|}{0.6261} & \multicolumn{1}{c|}{0.4660}          & \textbf{0.6391} \\ \hline
\end{tabular}
\end{table}
 \begin{figure*}[htbp]
	\centering
	\subfigure{
		\includegraphics[width=0.45\textwidth]{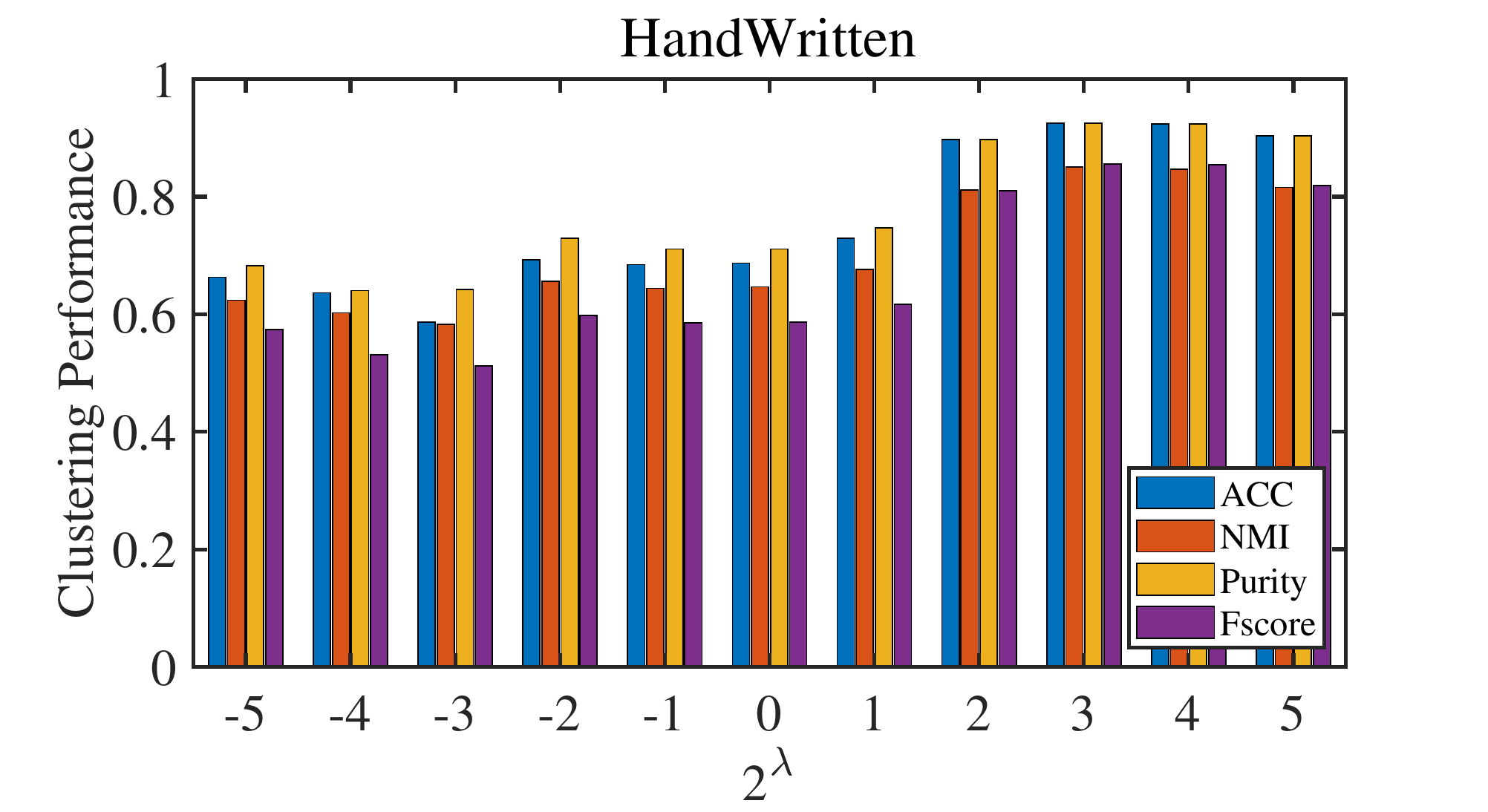}}
	\hspace{-0.2cm}
	\subfigure{
		\includegraphics[width=0.45\textwidth]{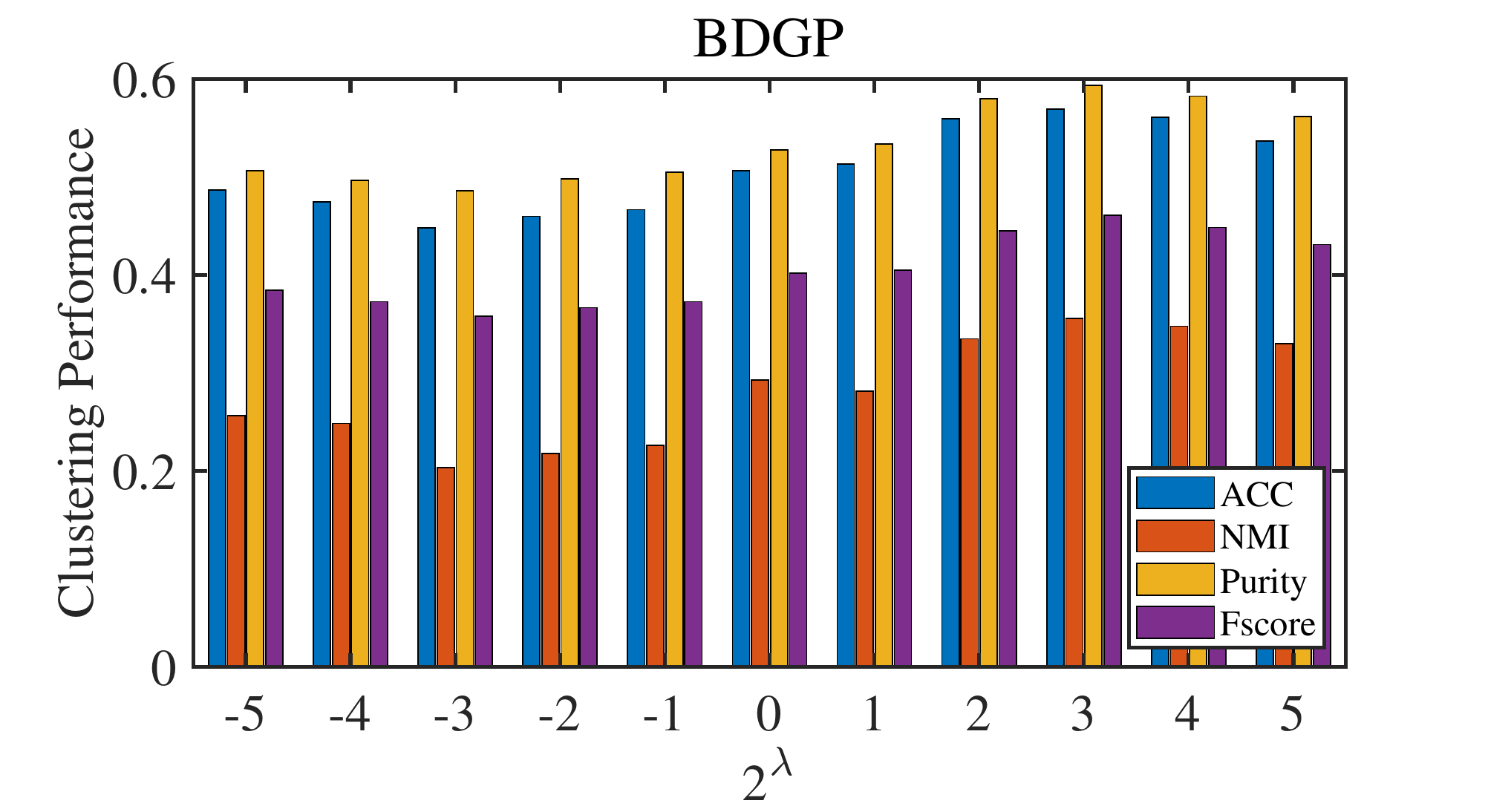}}
	\hspace{-0.2cm}
	\subfigure{
		\includegraphics[width=0.45\textwidth]{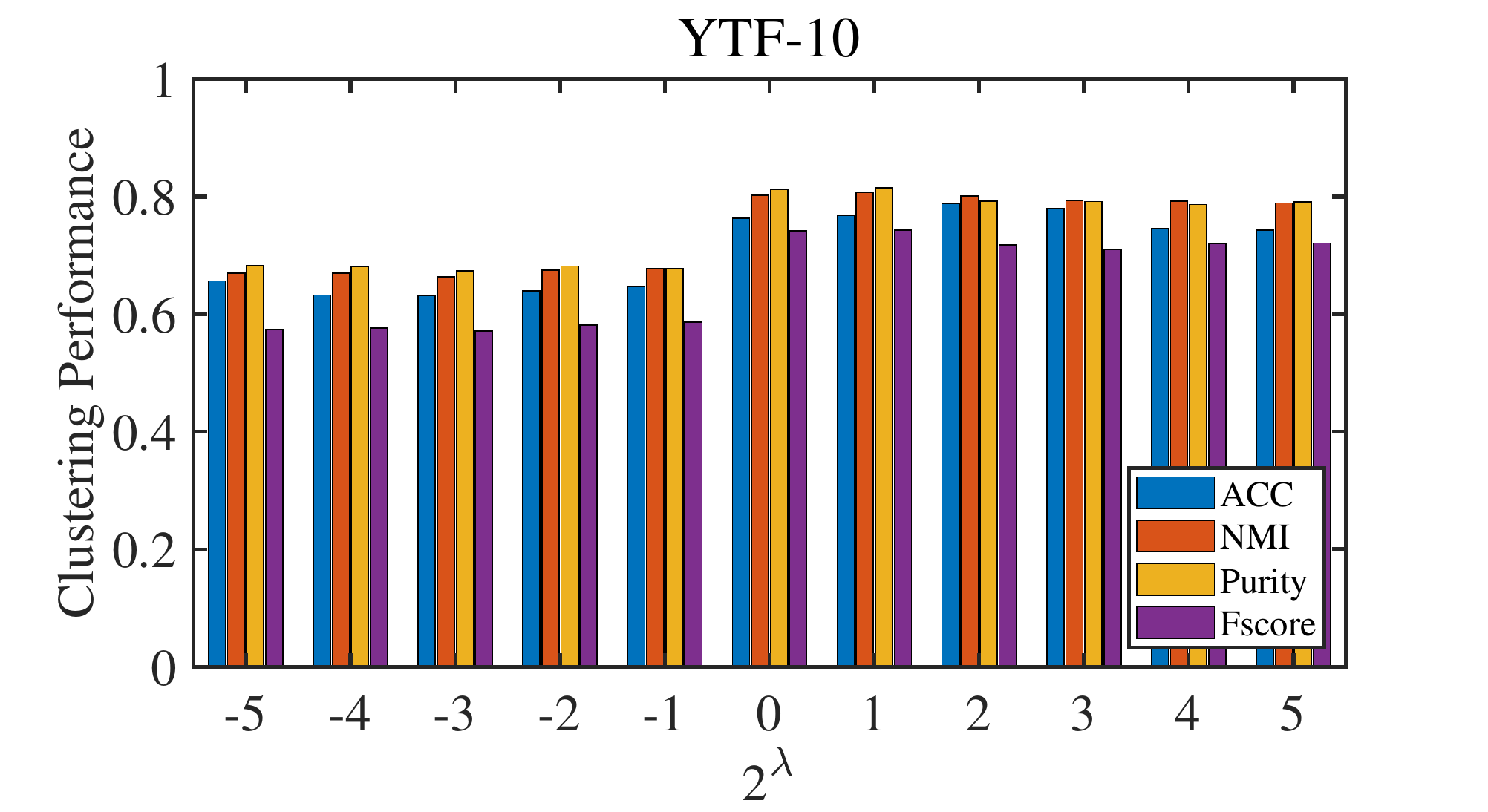}}
	\hspace{-0.2cm}
	\subfigure{
		\includegraphics[width=0.45\textwidth]{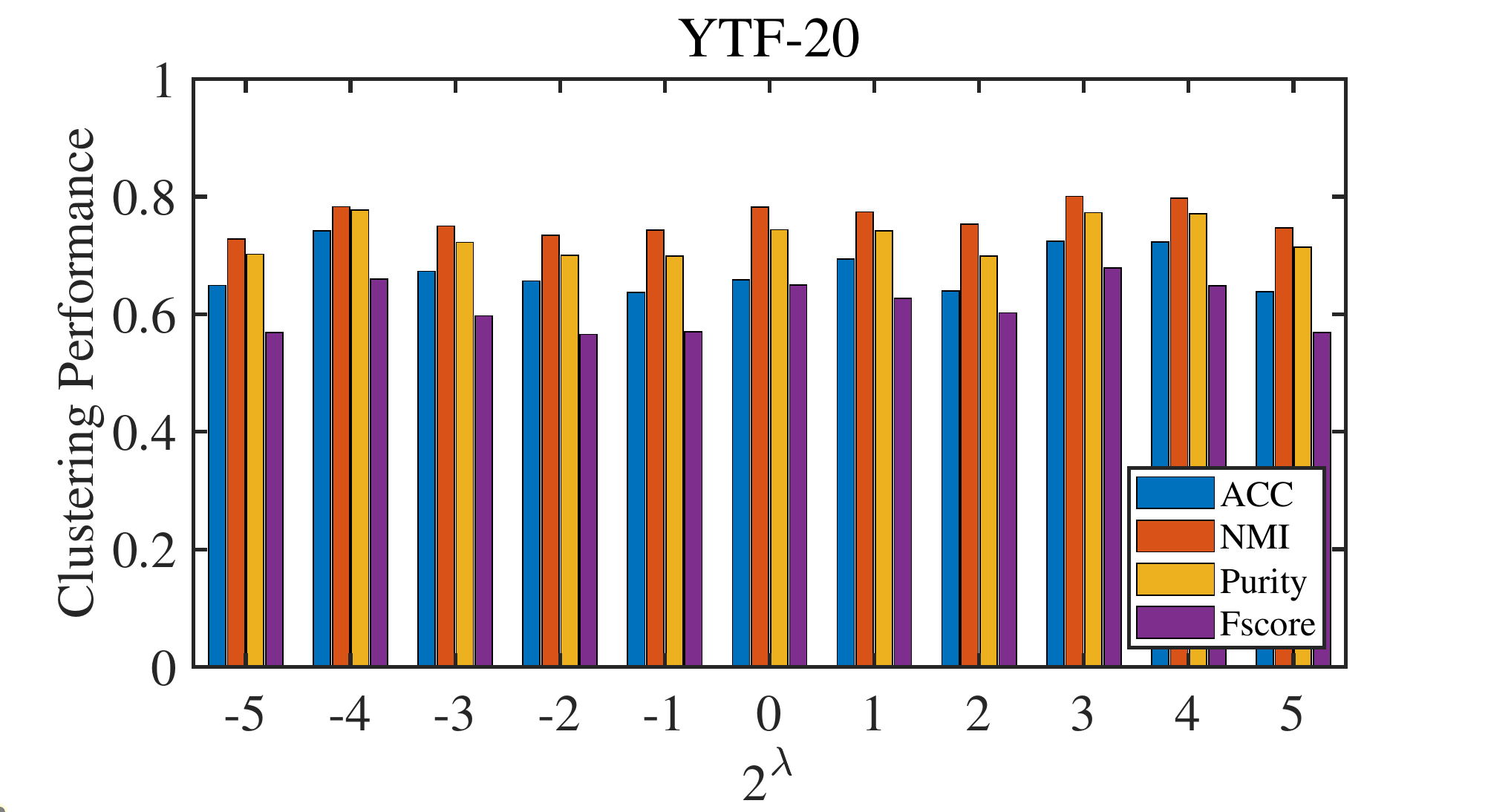}}
	\hspace{-0.2cm}

	\caption{The sensitivity of OMVCDR with the variation of $\lambda$ on four benchmark datasets, including HandWritten, BDGP,  YTF-10, and YTF-20, respectively.}
\label{fig_lambda}
\end{figure*}
Overall, our proposed method outperforms some state-of-the-art in terms of four clustering metrics, and we expect its effectiveness and efficiency will be considered for practical applications. In addition, we attribute the success of OMVCDR to reduced information loss. Compared with FMR, SMVSC, FPMVC, LMVSC and AWMVC, the interaction between consensus representation and clustering cluster is achieved. Contrasted with SFMC and OPMC, the diverse latent spaces of OMVCDR guides the representation learning with high quality.

\subsection{Running Time Comparison}
To demonstrate the efficiency of the proposed algorithm, we record the execution time of OMVCDR and the compared methods on seven benchmark datasets in Fig. \ref{running_time}. From the figure, it is obtained that:
\begin{enumerate}
\item Unlike traditional MVC algorithms such as FMR, the burgeoning anchor-based (SMVSC, FPMVS, LMVSC) or MF-based algorithms (OPMC, AWMVC, OMVCDR) consume less time and space to handle multi-view data and are more qualified for large-scale tasks. For instance, FMR costs the most time on Flower17, HandWritten, and BDGP and suffers from out-of-memory errors on other datasets. On the contrary, MF-based algorithms consume less time and can handle large-scale data.
\item OMVCDR enjoys comparable complexity with the compared methods. Although OPMC and LMVSC have less running time, their simple way of representation learning guides an unpleasing clustering result. Therefore, the time burden of our consensus knowledge acquisition and clustering is worthwhile.
\end{enumerate}

\subsection{Ablation Study}
OMVCDR works by unifying representation learning and clustering into a unified framework. Meanwhile, the diverse latent spaces and direct utilization of comprehensive knowledge contribute to the success of our proposed method. To evaluate the critical components of our proposed algorithm, we conduct an ablation study with three compared algorithms: 1) By setting $p=1$ and $p=2$, we investigate the effectiveness of diverse latent spaces in our model, termed OMVC and OMVCDR-2. 2) By assigning each $\alpha_p$ with the same value, we explore the effect of discrimination on different latent spaces to OMVCDR, called OMVCDR-$\alpha$. The results are described in Table \ref{ablation_study}. From the table, we can obtain that:
\begin{enumerate}
\item By simply setting $p=1$, the algorithm evolves as a traditional MFMVC method that projects the data matrix into a fixed dimension. The clustering performance of it is inferior to OMVCDR-2 and OMVCDR. Consequently, the strategy of comprehensive knowledge gain is beneficial.
\item Compared with OMVCDR-$\alpha$, OMVCDR achieves better clustering performance. It indicates that the discriminatory information guides a hard partition matrix with higher quality.
\item Distinct from AWMVC, we straightforwardly employ the information matrices to perform clustering and attain better results. Thus, the proposed one-pass method is favorable to our framework. 
\end{enumerate}

All in all, the clustering performance descends when one of the critical components of OMVCDR is dropped out. Therefore, the success of our work is related to the mapping of data to multiple dimensions, the obtained discriminatory information, and the one-step framework. 
\subsection{Convergence of the Proposed Algorithm}
As proved above, our proposed algorithm is theoretically convergent to a local minimum. To verify it in practice and discuss the convergence rate, we plot the objective value of OMVCDR changes with iterations on four datasets, shown in Fig. \ref{fig_obj}. From the figure, we can conclude that the objective value decreases monotonously and converges in less than 40 iterations, validating the convergence analysis in Section \ref{convergence_section} experimentally.
\subsection{Parameter Sensitivity}
We introduce a hyper-parameter $\lambda$ to balance the weight between representation learning and clustering. To investigate the influence of the value of $\lambda$ on our model, we perform a grid search and plot the results in Fig. \ref{fig_lambda}. From the figure, we see that the clustering performance first increases and remains stable with the variation of $\lambda$, indicating that OMVCDR is insensitive to the hyper-parameter.

\section{Conclusion}

This paper proposes a novel matrix factorization-based method termed One-step Multi-view Clustering with Diverse Representation (OMVCDR). Distinct from existing large-scale multi-view clustering algorithms, OMVCDR projects data matrices into diverse latent spaces and directly utilizes comprehensive knowledge to get a discrete clustering partition matrix with discriminatory information. To solve the resultant problem, we develop a four-step alternating algorithm with proven convergence, both theoretical and experimental. The idea of diverse latent spaces is general and can be easily extended to other areas, such as anchor learning. Comprehensive experiments demonstrate the superior performance of OMVCDR, and the ablation study verifies its effectiveness in the critical components.


%

\ifCLASSOPTIONcompsoc
  \section*{Acknowledgments}
\else
  \section*{Acknowledgment}
\fi

This work is supported by the National Key R$\&$D Program of China (project no. 2022ZD0209103, 2020AAA0107100), and the National Natural Science Foundation of China (project no. 61922088).

\ifCLASSOPTIONcaptionsoff
  \newpage
\fi



%

%

\begin{IEEEbiography}[{\includegraphics[width=1in,height=1.10in,clip,keepaspectratio]{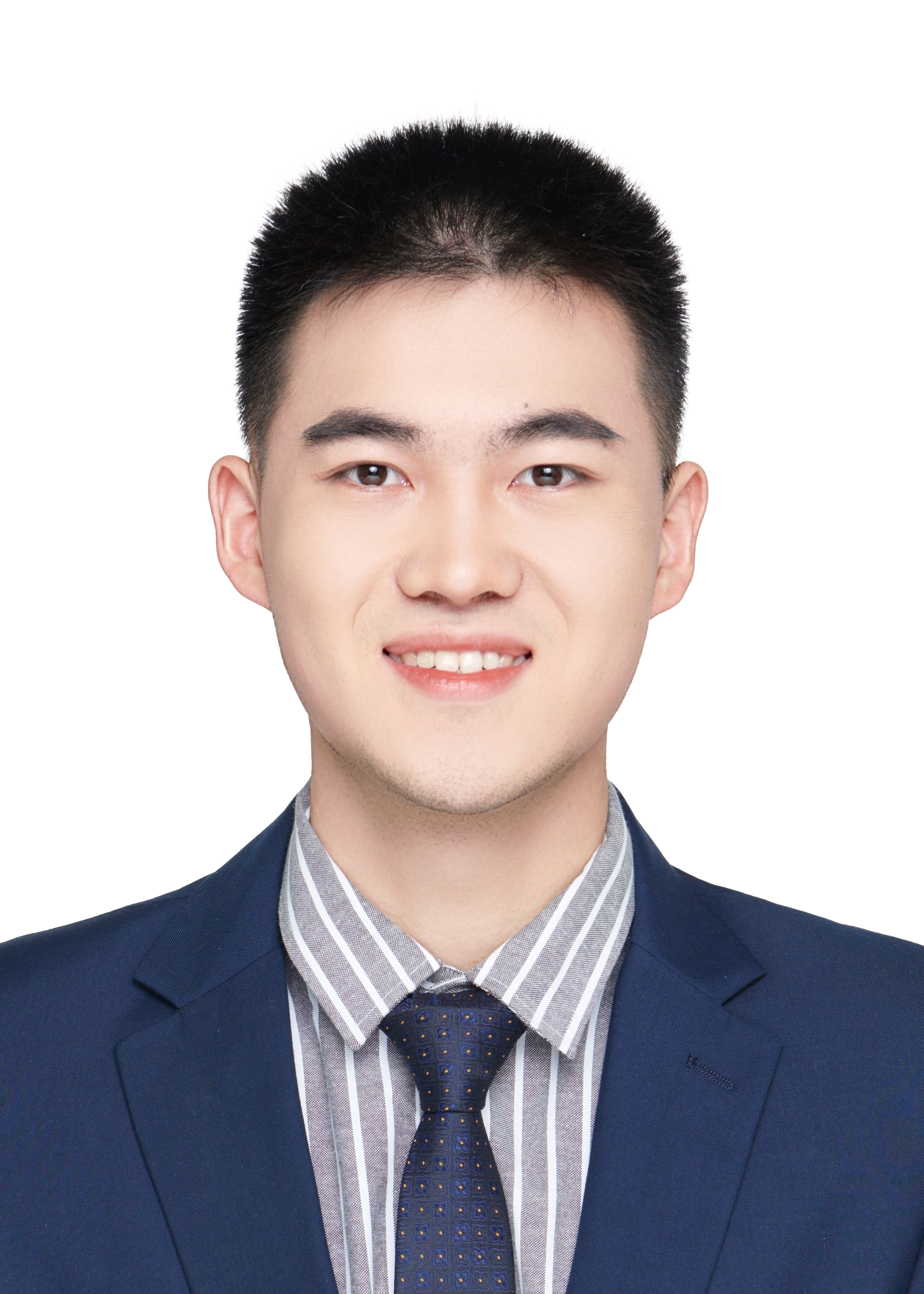}}]{Xinhang Wan} received the B.E degree in Computer Science and Technology from Northeastern University, Shenyang, China, in 2021. He is currently pursuing the Ph.D. degree with the National University of Defense Technology (NUDT), China.
 He has published papers in journals and conferences such as IEEE T-NNLS, ACM MM and AAAI. His current research interests include multi-view learning, incomplete multi-view clustering, and continual clustering.
\end{IEEEbiography}

\begin{IEEEbiography}[{\includegraphics[width=1in,height=1.10in,clip,keepaspectratio]{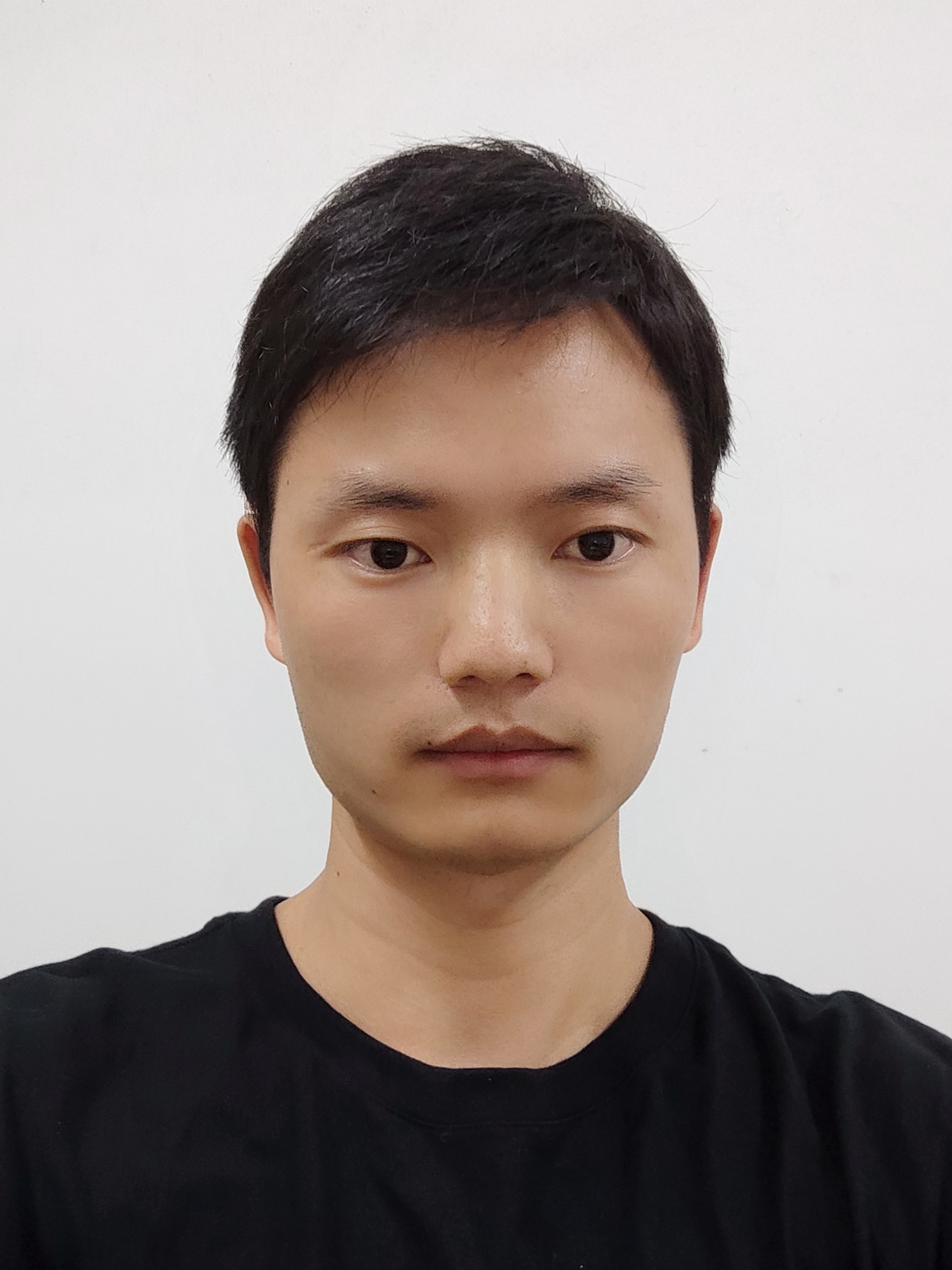}}]{Jiyuan Liu} received his PhD from National University of Defense Technology (NUDT), China, in 2022. He is now a lecturer with the College of Systems Engineering, NUDT. His current research interests include multi-view clustering, federated learning and anomaly detection. Dr. Liu has published papers in journals and conferences such as IEEE T-KDE, IEEE T-NNLS, ICML, NeurIPS, CVPR, ICCV, ACMMM, AAAI, IJCAI, etc. He serves as program committee member and reviewer on IEEE T-KDE, IEEE T-NNLS, ICML, NeurIPS, CVPR, ICCV, ACMMM, AAAI, IJCAI, etc. More information can be found at \url{https://liujiyuan13.github.io/}.
\end{IEEEbiography}

\begin{IEEEbiography}[{\includegraphics[width=1in,height=1.10in,clip,keepaspectratio]{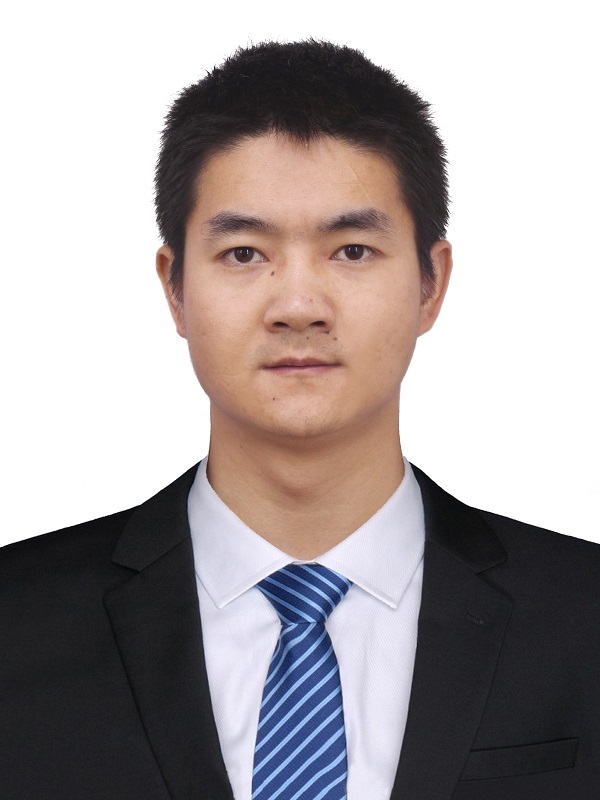}}]{Xinwang Liu} received his PhD degree from National University of Defense Technology (NUDT), China. He is now Professor of School of Computer, NUDT. His current research interests include kernel learning and unsupervised feature learning. Dr. Liu has published 60+ peer-reviewed papers, including those in highly regarded journals and conferences such as IEEE T-PAMI, IEEE T-KDE, IEEE T-IP, IEEE T-NNLS, IEEE T-MM, IEEE T-IFS, ICML, NeurIPS, ICCV, CVPR, AAAI, IJCAI, etc. He serves as the associated editor of Information Fusion Journal. More information can be found at \url{https://xinwangliu.github.io/}.
\end{IEEEbiography}

\begin{IEEEbiography}[{\includegraphics[width=1in,height=1.10in,clip,keepaspectratio]{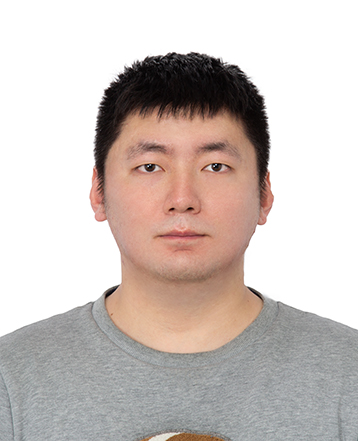}}]
{Siwei Wang} is currently pursuing the Ph.D. degree with the National University of Defense Technology (NUDT), China. He has published several papers and served as a PC Member/Reviewer
in top journals and conferences, such as IEEE
TRANSACTIONS ON KNOWLEDGE AND DATA ENGINEERING (TKDE), IEEE TRANSACTIONS ON NEURAL NETWORKS AND LEARNING SYSTEMS (TNNLS), IEEE TRANSACTIONS ON IMAGE PROCESSING (TIP), IEEE TRANSACTIONS ON CYBERNETICS (TCYB), IEEE TRANSACTIONS ON
MULTIMEDIA (TMM), ICML, CVPR, ECCV, ICCV, AAAI, and IJCAI. His current research interests include kernel learning, unsupervised multiple-view
learning, scalable clustering, and deep unsupervised learning.
\end{IEEEbiography}

\begin{IEEEbiography}[{\includegraphics[width=1in,height=1.10in,clip,keepaspectratio]{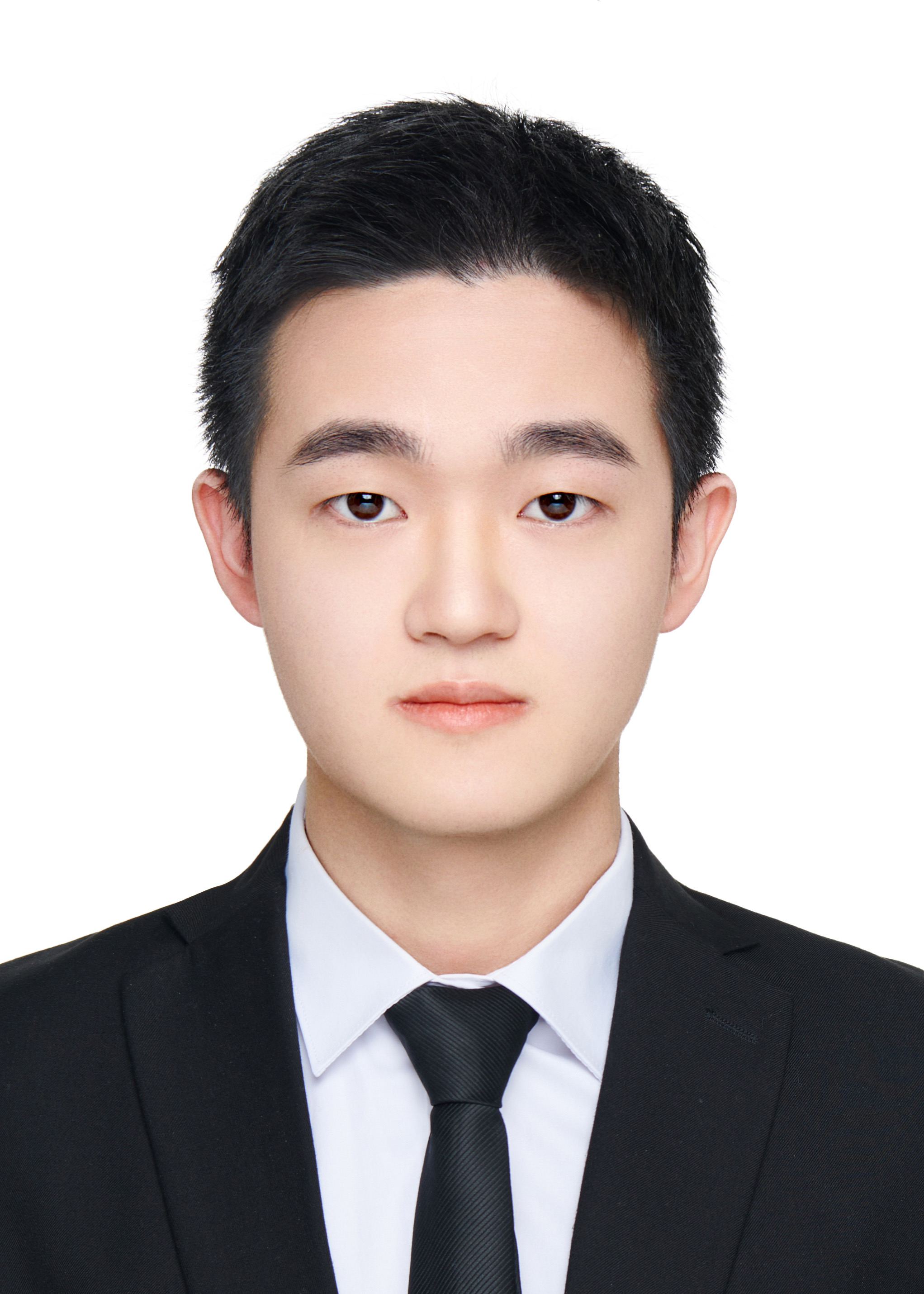}}]
{Yi Wen} is recommended for admission to the National University of Defense Technology (NUDT) as a master's student with  excellent grades and competition awards. He is working hard to pursue his master degree. His current research interests include multiple-view learning, scalable kernel k-means and graph representation learning.
\end{IEEEbiography}


\begin{IEEEbiography}[{\includegraphics[width=1in,height=1.10in,clip,keepaspectratio]{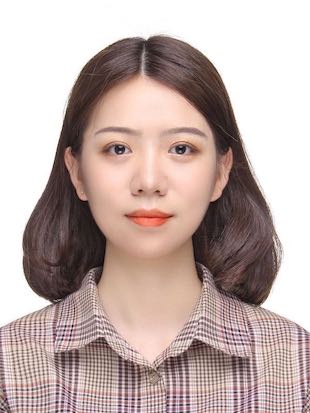}}]{Tianjiao Wan} graduated from Wuhan University in 2020 with a B.E. degree in software engineering. She is currently working toward a Doctor's degree with the National University of Defense Technology (NUDT), Changsha, China. Her current research interests include active learning, continual learning.
\end{IEEEbiography}

\begin{IEEEbiography}[{\includegraphics[width=1in,height=1.10in,clip,keepaspectratio]{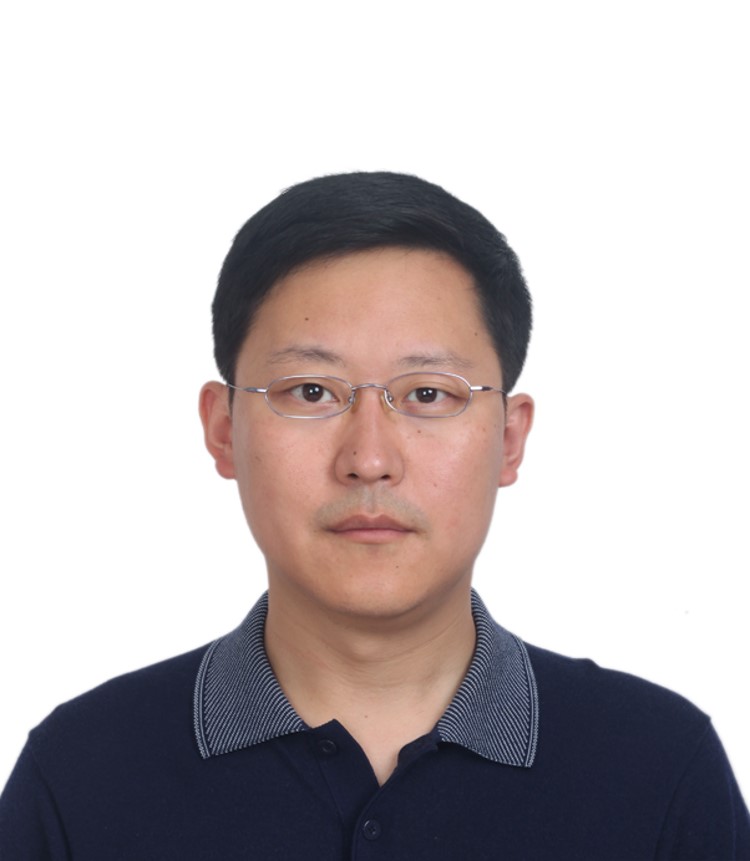}}]
{Li Shen} received the Ph.D. degree from the National University of Defense Technology (NUDT), Changsha, China, in 2003. He is currently a Professor with the School of Computer Science, NUDT. His current research interests include image super-resolution, machine learning, and performance optimization of machine learning systems. He has authored or coauthored 40 research papers, including
the IEEE TRANSACTIONS ON COMPUTERS, IEEE TRANSACTIONS ON PARALLEL AND DISTRIBUTED SYSTEMS (TPDS), Micro, IEEE International Symposium on High-Performance Computer Architecture (HPCA), and Design Automation Conference (DAC).
\end{IEEEbiography}

\begin{IEEEbiography}[{\includegraphics[width=1in,height=1.10in,clip,keepaspectratio]{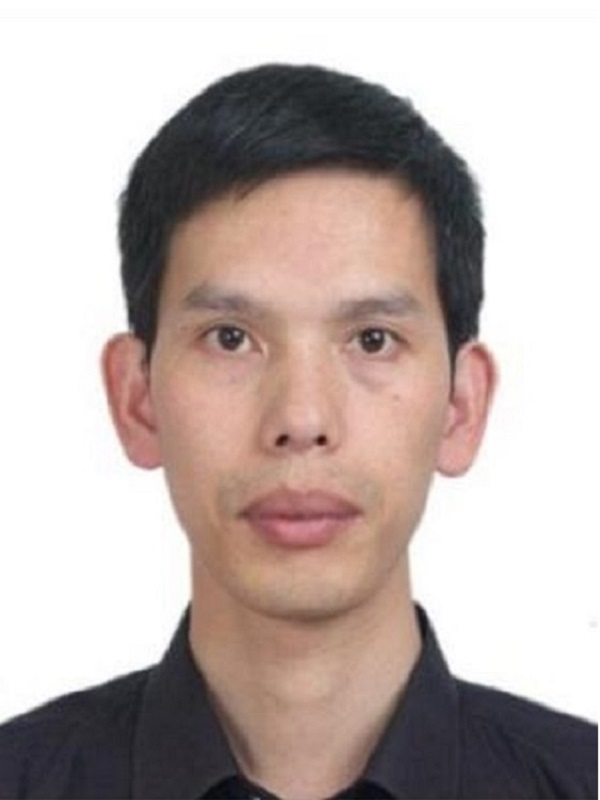}}]{En Zhu} received the Ph.D. degree from the National University of Defense Technology (NUDT), China. He is currently a Professor with the School of Computer Science, NUDT. He has published more than 60 peer-reviewed papers, including IEEE TRANSACTIONS ON CIRCUITS AND SYSTEMS FOR VIDEO TECHNOLOGY (TCSVT), IEEE TRANSACTIONS ON NEURAL NETWORKS AND LEARNING SYSTEMS (TNNLS), PR, AAAI, and IJCAI. His main research interests include pattern recognition, image processing, machine vision, and machine learning. He was awarded the China National Excellence Doctoral Dissertation.
\end{IEEEbiography}




\end{document}